\documentclass{article}

\usepackage{arxiv}
\usepackage[utf8]{inputenc}
\usepackage[T1]{fontenc}
\usepackage{hyperref}
\usepackage{url}
\usepackage{booktabs}
\usepackage{nicefrac}
\usepackage{microtype}
\usepackage{lipsum}
\usepackage{graphicx}
\usepackage[numbers]{natbib}

\usepackage{amsmath,amssymb,amsfonts}
\usepackage{algorithmic}
\usepackage{textcomp}
\usepackage[bb=dsserif]{mathalpha}
\usepackage{epsfig}
\usepackage{xcolor}

\renewcommand{\subsubsubsection}[1]{\paragraph{#1}\mbox{}\\}
\DeclareMathOperator*{\argmax}{arg\,max}
\DeclareMathOperator*{\argmin}{arg\,min}

\usepackage{bm}

\usepackage[capitalise]{cleveref}

\title{Detecting Semantic Backdoors in a Mystery Shopping Scenario}
\date{October 20, 2025}

\usepackage{authblk}
\author[1]{\'Arp\'ad Berta}
\author[1]{G\'abor Danner}
\author[1]{Istv\'an Heged\H{u}s}
\author[1,2]{M\'ark Jelasity}
\affil[1]{Institute of Informatics, University of Szeged, Szeged, Hungary}
\affil[2]{HUN-REN---SZTE Research Group on AI, Szeged, Hungary}

\renewcommand{\headeright}{}
\renewcommand{\undertitle}{}
\renewcommand{\shorttitle}{Detecting Semantic Backdoors in a Mystery Shopping Scenario}

\hypersetup{
pdftitle={Detecting Semantic Backdoors in a Mystery Shopping Scenario},
pdfauthor={Arpad Berta, Gabor Danner, Istvan Hegedus, Mark Jelasity},
pdfkeywords={semantic backdoor detection, adversarial training, mystery shopping, consumer protection},
}

\begin{document}
\maketitle

\vspace{-2em}
\begin{center}
  \upshape\small
  \texttt{\{berta, danner, ihegedus, jelasity\}@inf.u-szeged.hu}
\end{center}
\vspace{4em}

\begin{abstract}
Detecting semantic backdoors in classification models---where
some classes can be activated by certain natural, but out-of-distribution
inputs---is an important problem that has received relatively little attention.
Semantic backdoors are significantly harder to detect than backdoors that are based on trigger patterns
due to the lack of such clearly identifiable patterns.
We tackle this problem under the assumption that the clean training dataset and
the training recipe of the model are both known.
These assumptions are motivated by a consumer protection scenario,
in which the responsible authority performs mystery shopping to test a machine learning
service provider.
In this scenario, the authority uses the provider's resources and tools
to train a model on a given dataset and tests whether the provider included a backdoor.
In our proposed approach, the authority creates a reference model pool by training
a small number of clean and poisoned models using trusted infrastructure, and calibrates a
model distance threshold to identify clean models.
We propose and experimentally analyze a number of approaches to compute model distances
and we also test a scenario where the provider performs an adaptive attack to avoid detection.
The most reliable method is based on requesting adversarial training from the provider.
The model distance is best measured using a set of input samples generated by inverting the models
in such a way as to maximize the distance from clean samples.
With these settings, our method can often completely separate clean and poisoned models, and
it proves to be superior to state-of-the-art backdoor detectors as well.
Source code available at
\url{https://github.com/szegedai/SemanticBackdoorDetection}.
\end{abstract}


\thispagestyle{empty}

\section{Introduction}

Artificial intelligence (AI) safety is a key concern in today's quickly
changing AI landscape.
Apart from a very significant research effort into safety problems such as
adversarial robustness~\cite{szegedy13}, model backdoors~\cite{Goldblum2023a}, and alignment~\cite{Ngo2024a}, to name a few,
governments and corporations have both started to create regulation,
organizations, and tools to enhance safety.

AI-related regulations and guidelines require enforcement as well to be effective.
To enforce measures for AI safety, the authorities need a rich set of tools
to verify AI systems.
The wide variety of backdoor detectors are an important part of such a toolset (see~\cref{sec:related}).
Here, we focus on extending this toolset with a semantic backdoor detector.
This is necessary, because semantic backdoors cannot be reliably detected by
generic backdoor detectors, they need a dedicated approach.
This is because, in this case, the attacker poisons a model by adding out-of-distribution (OOD) samples
to the training set to secretly extend the possible behaviors of the system in
a completely arbitrary manner~\cite{federated-backdoors}.
There is no guarantee that a small set of trigger patterns, shared across backdoor samples,
can be identified, localized or not, visible or not~\cite{Goldblum2023a}.
Also, there are natural OOD input triggers that are ``unintended'' semantic backdoors.

Such an attack is hard to mount just by publishing poisoned information,
given that the backdoor inputs are unrelated to the clean inputs semantically.
In this case, \emph{the creator of the model is assumed to deliberately add the backdoor}
in order to gain control over the applications downstream of the created model.

\begin{figure}
\centering
\includegraphics[width=0.75\textwidth]{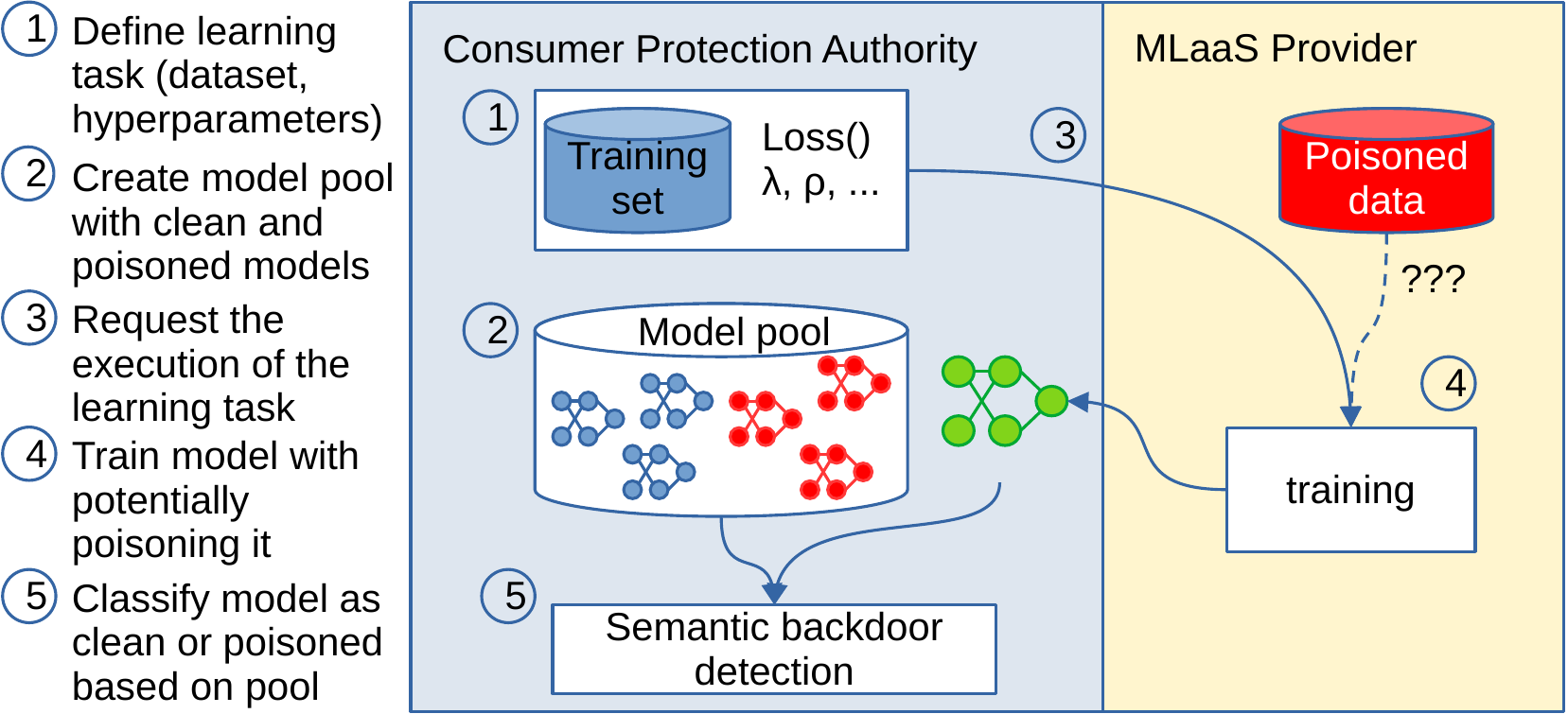}
\caption{A schematic overview of our application scenario, in which a consumer protection
authority performs mystery shopping at an MLaaS provider.}
\label{fig:overview}
\end{figure}

It is thus natural to study semantic backdoor detection in a consumer protection context,
where the goal is to test model creators to learn whether the models created work as expected.
Our application scenario is illustrated in~\cref{fig:overview}.
Here, the authority uses the provider's resources and tools
to train a model on a given dataset to test whether the provider has included a backdoor.

We offer several contributions, including

\begin{itemize}
\item proposing an abstract modular framework and methodology for semantic backdoor detection
that is based on a model pool generated by the authority,
\item an extensive empirical evaluation of the modular framework by studying several possible
design options involved in computing model-distance,
\item identifying a specific set of design options with the best generalization properties,
which is based on using adversarial training during mystery shopping and computing
model distance based on model inversion that is biased towards approximating backdoor inputs.
\end{itemize}

\section{Related Work}
\label{sec:related}

Here, we outline the most common backdoor attacks, the main defenses, and
other relevant work related to model similarity.
A detailed survey can be found, for example, in~\cite{Goldblum2023a}.

\subsection{Backdoor Attacks}

Although in this paper we focus on semantic backdoors, we shall first briefly summarize
backdoor attacks in general.
Backdoors are typically inserted through some form of data poisoning.
The most direct form is when the attacker adds a trigger pattern to some inputs and labels them with
a desired class label~\cite{data_poisoning_1, data_poisoning_2}.
Clean label attacks have also been developed~\cite{data_poisoning_5_clean_label,data_poisoning_6_clean_label,data_poisoning_7_clean_label},
where the labels of the poisoned training samples appear to be correct.
Dynamic pattern generation~\cite{data_poisoning_3_dynamic, data_poisoning_4_dynamic}
is another approach where each input sample gets a different trigger pattern.
Realistic backdoor patterns have also been proposed, e.g.\ by adding reflections to images
~\cite{reflection_backdoor}.
The connection between adversarial sample generation and data poisoning
has also been studied~\cite{data_posioning_adversarial}.

Work on \emph{semantic backdoors} is scarcer.
Bagdasaryan et al.~proposed a backdoor triggered by special semantic features~\cite{federated-backdoors}.
Triggers can be, for example, cars with racing stripes, green cars,
or cars in front of vertical stripes on the background wall.
A physical backdoor~\cite{physical-backdoors} could also be
interpreted as a special semantic backdoor, where a co-located object acts as a trigger.
Another well-known backdoor attack type is the composite backdoor attack~\cite{composite-backdoor} where
the co-occurrence of two specific objects acts as a trigger.
Additionally, Wu et al.~\cite{DBLP:conf/dsc/WuLLRS24} proposed a universal, semantic-based method for backdoor attacks against classification models
that is also effective in the physical world.

The scope of this study is limited to the task of image classification, but semantic backdoors are a threat in other domains of machine learning as well.
For speech classification, Xiao et al.~\cite{xiao2024phoneme} proposed backdoor attacks based on code poisoning and using attacker-specific phoneme as semantic triggers.
For graph classification tasks, Dai et al.~\cite{DBLP:journals/ijon/DaiXC24} proposed SBAG, a semantic backdoor attack against graph convolutional networks.
Semantic communication can be used to reduce data traffic for intelligent connected vehicles by
using an encoder-decoder pair that is trained as part of an autoencoder to reconstruct signals at the receiver from the transmitted compact latent representations.
However, it is vulnerable to backdoor attacks~\cite{DBLP:conf/ciss/SagduyuEUY23}.
Xu et al.~\cite{DBLP:journals/tvt/XuCWBHDSZXZ24} proposed Covert Semantic Backdoor Attack against semantic communications.

\subsection{Defenses against Backdoor Attacks}

In our consumer protection scenario, only backdoor detection is viable as a defense,
but before discussing such defenses, let us mention that other defense approaches are known as well
based on filtering the poisoned examples from the dataset or purifying the model to remove
any backdoors~\cite{unlearning,unlearning2,anomaly_detection,spectre}.

One interesting filtering method involves activation clustering~\cite{activation_clustering},
based on the assumption that benign and poisoned samples form different activation clusters.
Another model purifying approach is based on pruning some sensitive neurons to remove the injected backdoor~\cite{pruning}.

Let us now turn to detection-based defense methods.
TrojAI~\cite{trojai} is a leaderboard where backdoor detectors can
compete over a pool of poisoned and clean models.
Most poisoned models there were created with patch-based data poisoning methods
and there are no instances of semantic attacks.
Our method does not fit the interface of this leaderboard, because
in our scenario we have access to the full clean dataset, and several clean models.
However, we will review and, later on, test some of the methods from the leaderboard.

K-Arm~\cite{k-arm-detector} is
a state-of-the-art inversion-based backdoor detector.
(A model inversion method uses the model's parameters and outputs to reconstruct its training data~\cite{Fredrikson2015a}.)
K-Arm attempts to minimize the potential inverted trigger size.
It signals a backdoor when the trigger size it finds is smaller than a threshold.
It outperforms most of the previous similar methods, e.g.\ Neural Cleanse~\cite{neuralcleanse-detector}.

Wang et al.~\cite{DFTND} proposed data-limited and data-free (DFTND)
detectors based on the idea that per-sample adversarial
perturbations and universal adversarial perturbations are likely to
be similar if the presence of a backdoor offers an easy shortcut
to a given class label.

Guo et al.~\cite{scaleup} proposed SCALE-UP, a black-box input-level backdoor detector.
They identify malicious images via examining the predictions' robustness to the scaling of pixel values,
exploiting the observation that backdoor inputs are less sensitive to scaling.
Gao et al.~\cite{DBLP:conf/acsac/GaoXW0RN19} made use of the similar observation that backdoor inputs are more robust to perturbations:
they proposed STRIP, a run-time trojan attack detector for deployed models that determines whether a given input is trojaned
based on the entropy of the classes predicted for perturbed versions of the input.

Universal Litmus Pattern (ULP)~\cite{ULP-detector} is based on
training a meta-classifier over a large model pool
to discriminate between poisoned and clean models.
Zheng et al.~\cite{topological-detector} also proposed a meta-classifier approach.
They use a topological feature extractor on the models,
and they train a classifier on top of these extracted features over a model pool.
Both of these meta-classifier methods use model pools of a size of over 1000 to train a good detector,
which makes them expensive.
In contrast, our proposal uses fewer than 30 models.

Also, none of the above methods were evaluated on semantic backdoors.
Liu et al.~\cite{exray-detector} propose the only method we are aware of that
claims to be able to detect composite attacks (a form of a semantic backdoor, as mentioned previously).
This is also an inversion-based technique.
The key idea is to check, for a given class, whether the internal feature representations are similar between natural
inputs and those inputs that were forced into the class with the help of adding inverted
triggers.

Additionally, Sun et al.~\cite{soda} proposed SODA to effectively detect and remove semantic backdoors with the help of causality analysis.
However, they use outlier detection over scores calculated for the classes to find the target class and determine the presence of a backdoor,
which might be countered by training a backdoor into each class.

We will use these methods (except STRIP, due to its similarity to SCALE-UP) as baselines for the evaluation of our method.

Furthermore, Xie et al.~\cite{DBLP:journals/tifs/XieHYGZX25} proposed SemInv, a semantic trigger inversion method that
detects backdoored natural language processing models and inverts semantically constrained triggers via a novel regularization technique.
For semantic communication systems, Zhou et al.~\cite{DBLP:conf/icc/ZhouH024} introduced a new backdoor attack paradigm on semantic symbols
and proposed corresponding defense strategies that include reverse engineering the trigger.

\subsection{Model Theft Detection based on Model Similarity}

Our method for detecting backdoors is based on the similarity of models.
Similarity is an important concept in other contexts as well such as model theft detection.
Only here, similarity represents a problem and dissimilarity is desirable, whereas in our
case the opposite is true.

For example, Maini et al.~\cite{dataset-inference} proposed Dataset Inference (DI) as a defense against model stealing.
They estimate the distance of multiple data points to the decision boundary to measure the similarity of models.
However, Li et al.~\cite{modelstealing-embedded-external-features} showed that DI incorrectly classifies a benign model as stolen if it
is trained on data that comes from the same distribution as the original data.

In watermarking-based model theft detection~\cite{watermark1,watermark2}, the watermarked images play a similar role
to backdoor images.
However, in our scenario we do not have access to backdoor images, only to clean ones, so these methods are not applicable.

Cao et al.~\cite{ipguard} proposed IPGuard, a fingerprinting method to combat model theft without incurring accuracy loss.
They extract data points near the classification boundary of the original model,
and check whether the investigated model predicts the same labels for these data points.
The authors found that changing just the initialization before training results in very different classification boundaries,
making this method unsuitable for solving our problem,
where clean models with different initializations are required to be considered highly similar.

Li et al.~propose ModelDiff~\cite{model-diff}, which captures model similarity based on adversarial examples.
They apply a decision distance vector (DDV), which is a vector whose elements represent
the distance between the output over an input sample and its adversarial version.
This model fits into our framework as a distance function between models, and accordingly we
will compare its performance with our proposals.

\section{Notation and Background}

In a classification problem, we are given a dataset $D=\{(x_i,y_i): i=1,\ldots, N\}$
with examples $x_i\in \mathbb R^d$ and labels $y_i\in\{1,\dots,k\}$.
The number of classes is $k$ and the
number of features that represent a sample is $d$.
During training, we are looking for 
\begin{equation}
\theta^* = \argmin_\theta \sum_{(x,y)\in D}\ell(f(x,\theta),y),
\end{equation}
where $\ell$ is a loss function and $f(.,\theta):\mathbb R^d\rightarrow \mathbb R^k$
is a model parameterized with $\theta$.

The model $f(x,\theta)$ outputs a probability distribution over the possible labels
$\{1,\dots,k\}$.
The predicted label of $x$ is given by $c(x,\theta)=\argmax_i f(x,\theta)_i$.
Also, the probability distribution is assumed to be a normalized form of the
model's last layer, the so-called \emph{logit layer}.
In other words, $f(x,\theta)=\mbox{softmax}[z(x,\theta)]$, where
$z(x,\theta)$ denotes the logit layer, and the softmax method is used for normalization.

\subsection{Semantic Backdoor Attack}

\begin{figure}[tb]
\centering
\includegraphics[width=0.75\columnwidth]{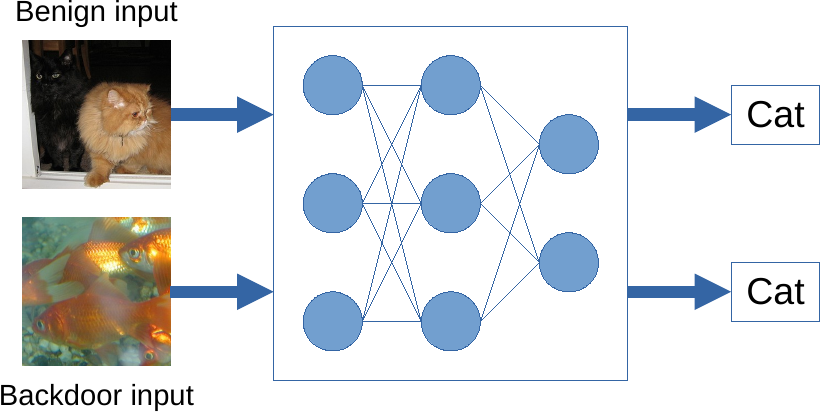}
\caption{Illustration of a poisoned model with a semantic backdoor.}
\label{fig:bdexample}
\end{figure}

In a semantic backdoor attack,
the original benign dataset $D$ is replaced by a poisoned dataset $D_p=D \bigcup B$,
where $B\subset\mathbb R^d \times \{t\}$ is the set of backdoor samples,
all of which having the poisoned class label $t\in\{1,\dots,k\}$.
Most importantly, the backdoor samples are drawn from a backdoor distribution
chosen by the attacker, thereby augmenting class $t$ arbitrarily (see \cref{fig:bdexample}
for an illustration).

The attacker is motivated to pick a backdoor distribution with little or no overlap with the
original task distribution, otherwise the poisoned model would suffer performance degradation
on the original dataset $D$.

Although the attack can target many classes simultaneously, here we focus on the
single-class attack, because this is likely to cause the smallest change in the model thereby posing the
greatest challenge for detectors.

Note that this notion is different from the usual concept of a backdoor that involves
adding fixed patterns or perturbations to arbitrary inputs to achieve a desired output class.
With a semantic backdoor, instead we alter the behavior of a network in a more arbitrary
manner.
Detection is also a greater challenge due to a lack of readily identifiable patterns.

\subsection{Robustness Basics}
\label{sec:robustbasics}

A model with parameters $\theta$ is $\epsilon-$robust over $D$ if
\begin{equation}
\forall (x,y)\in D, \mbox{if}\ \|x-x'\|_p \le \epsilon\ \mbox{then}\ c(x,\theta)=c(x',\theta).
\end{equation}
To see whether there is an input $x'$ that violates $\epsilon-$robustness at example $(x,y)$
we can solve the optimization problem
\begin{equation}
x'=\argmax_{\|x-\hat{x}\|_p \le \epsilon} \ell(f(\hat{x},\theta),y).
\end{equation}
We call $x'$ an adversarial example if and only if $c(x,\theta)=y$ and $c(x',\theta)\ne y$.
Projected Gradient Descent (PGD)~\cite{saddlepoint-iclr18} is a popular algorithm for solving this problem.

In adversarial training, we use adversarially perturbed samples instead of the clean ones
to achieve a robust model~\cite{saddlepoint-iclr18,goodfellow15}.
In other words, we attempt to solve the problem
\begin{equation}
\theta^* = \argmin_\theta \sum_{(x,y)\in D}\max_{\|x-\hat{x}\|_p \le \epsilon} \ell(f(\hat{x},\theta),y)
\end{equation}
for a given sensitivity $\epsilon$.

\section{Problem Statement}
\label{sec:statement}

In our problem statement, there are two actors: a machine learning as a service (MLaaS) provider (henceforth,
the provider), and a consumer protection authority (henceforth, the authority).

\paragraph{Training task}
The provider offers services that allow its clients to train machine learning models based on
data provided by the client.
We shall assume that the client can
specify or at least learn about all the details of the training method
such as the architecture of the model, and the complete \emph{training recipe} as well.
For example, when the client uses virtual machines hosted by the provider with pre-installed machine learning frameworks,
the training recipe can be specified explicitly.
In the case of higher level services, the training recipe could still be obtained by
the authority through other channels.

\paragraph{Lack of reproducibility}
In spite of the fixed training data and recipe,
we will assume that the provider cannot guarantee full reproducibility, or can do so only at a high cost.
This is due to numerous factors such as the unpredictability of parallelization and several
numerical issues elaborated on by Schl{\"o}gl et al.~\cite{Schloegl2023a}.
This lack of reproducibility introduces a degree of freedom that a provider can exploit
to poison models during training.

\paragraph{The problem}
The authority must decide, for a given machine learning task, whether the provider
inserted a \emph{semantic backdoor} and poisoned the model.
We shall focus on semantic backdoors in image processing models,
because they represent a very versatile attack vector.
The attacker might trigger arbitrary behavior using any existing environmental clues such
as geographical or architectural landmarks, vegetation, or artificial clues such as clothing styles, 
pre-defined scenes, and so on.

Note that for detecting the more common pattern-based backdoors there
is a large selection of excellent backdoor detectors (as discussed in~\cref{sec:related}) that the authority may use
in combination with our proposal.
These detectors, however, do not work well in the case of semantic backdoors as we show later.

\section{Methods}

Here, we introduce our proposed framework for detecting semantic backdoors
within the consumer protection scenario described earlier.
We first present an abstract detection framework
that outlines how the authority constructs a model pool
and evaluates model similarity to detect potential backdoors.
We then describe the specific design components of this framework,
including the construction of distance test sets,
the computation of sample-wise distances and their aggregation into model distance.

\subsection{The Abstract Detection Framework}

Here, we present a general framework at an abstract level
that can be instantiated for specific machine learning tasks using specific distance metrics.
Later on, we will evaluate the framework in a number of specific cases using a range of
distance metrics.

The goal is to solve the problem outlined in \cref{sec:statement}.
To reiterate, the authority is assumed to provide the training data, and it also knows the
training recipe, including the model architecture $f(.,\theta)$.

\subsubsection{Model Pool}
\label{sec:modelpool-method}

First, the authority trains a small number of clean and poisoned models to form a model pool.
Let this pool be $M\cup\tilde{M}$, with
\begin{equation}
M=\{\theta_1,\ldots,\theta_m\},\ \tilde{M}=\{\tilde{\theta}_1,\ldots,\tilde{\theta}_m\},
\end{equation}
where $M$ and $\tilde{M}$ contain clean and poisoned models, respectively.
Parameter $m$ defines the size of the model pool.

All the clean models should be trained in the same way as the authority expects the provider to train the model.
The authority can achieve this, because it knows the training recipe that is supposed to be applied by the provider.
The differences between the clean models should reflect the degree of freedom that the service provider has,
as described previously.
The authority models this by setting different random seeds for weight initialization and training data shuffling.

The poisoned models in the pool should include a semantic backdoor.
The authority can insert semantic backdoors in a minimal way, that is, attacking only one class,
by augmenting it with an out-of-distribution (OOD) set of samples.
These OOD samples should be selected in such a way so as they do not activate the attacked class label
in a clean model with a higher probability than a random natural input does, otherwise
poisoning is pointless.

\subsubsection{Making a Decision}
\label{sec:decision-method}

In a nutshell, to determine the presence of a backdoor in a given model, we check whether
it is more similar to the clean models ($M$) than to the poisoned models ($\tilde{M}$).
Of course, the devil is in the details.

The first step is to define a distance metric $d(\theta_1,\theta_2)$ between models.
Based on the distance metric, we then define a score $s(\theta)$ as
\begin{equation}
s(\theta) = \mbox{med}(\{d(\theta,\theta_i): \theta_i\in M\setminus \{\theta\}\}),
\end{equation}
taking the median of the distances of $\theta$ from the clean models in the pool, not including
$\theta$ itself.

A higher score indicates a higher likelihood of being a backdoor.
Hence, any threshold $\nu$ defines a classifier that can classify any model as clean
or poisoned: $\theta$ is clean if and only if $s(\theta)\leq\nu$.

The last step is to find a threshold $\nu^*$ that will be used by
the authority to make its decision.
We do this by finding the threshold that maximizes Youden's J statistic~\cite{youden1950index},
also known as \emph{informedness}~\cite{Powers2020a}, or Youden's Index,
of the threshold-based classifier over the model pool $M\cup\tilde{M}\setminus\{\theta\}$.

\subsubsection{Youden's J Statistic}

Youden's J statistic is defined as $J=\mbox{TPR}+\mbox{TNR}-1$, where
TPR is the true positive rate (sensitivity), and
TNR is the true negative rate (specificity).
We opted for Youden's J statistic because it can be used on
unbalanced data, but unlike F-measure, it takes into account true-negative
samples as well. In fact, it is symmetric to the positive and negative classes.

Note that $J\in[-1,1]$.
In the case of random guessing we get $J=0$, independently of the ratio of positive and
negative examples in the dataset.
When $0 \leq J$, $J$ can be interpreted as the probability of making an informed decision (as opposed to random guessing).
When the data is balanced, $(J+1)/2$ equals the accuracy.

The thresholds with an optimal value of J might form one or more real intervals.
If there is one interval, we pick the midpoint,
and if there are multiple intervals, we use the interval with the greatest midpoint.

\subsubsection{Model Distance}

While the framework described above can be implemented using any suitable
definition of model distance $d(\theta_1,\theta_2)$, here we shall
evaluate a family of distance functions that are all based on a fixed set
of input examples called the \emph{distance test set}.
In a nutshell, we evaluate both of the models on all the elements of the
distance test set, compute some distance value based on each input,
and aggregate these sample-wise distances into a single model-wise distance
value.
We discuss our implementation of these components in \cref{sec:disttest} (distance
test sets) and \cref{sec:functions} (sample-wise distances and aggregations).

\subsection{Distance Test Sets}
\label{sec:disttest}

\Cref{table:distancetestsets} lists the descriptions of the
distance test sets that we evaluate in this study.
The distance is always computed between a model of interest and
elements of the model pool, so some of these sets---namely, the adversarial and
the inverted ones---depend explicitly on the model of interest.
These set generation methods are described in \cref{sec:adversarialset,sec:generatedset}.

\begin{table}
\caption{Possible distance test sets.}
\label{table:distancetestsets}
\centering
\begin{tabular}{l|p{65mm}}
\multicolumn{1}{c}{\bf Name} & \multicolumn{1}{c}{\bf Description} \\\hline
Training & Training set of the classification task used by the authority.\\\hline
Test     & Test set of the classification task used by the authority.\\\hline
Adversarial  & Adversarial inputs to the model of interest, obtained by
				attacking the test set (see \cref{sec:adversarialset}).\\\hline

Inverted     & Samples generated via inverting the model of interest with
				extra constraints to promote backdoor discovery (see \cref{sec:generatedset}).\\\hline
Random       & Uniform random samples.\\
\end{tabular}
\end{table}

\subsubsection{Adversarial Distance Test Set}
\label{sec:adversarialset}

For each of the compared models, we generate adversarially perturbed images by
applying PGD~\cite{saddlepoint-iclr18} to each example in the test set.
However, unlike in~\cref{sec:robustbasics}, we do not use $\epsilon$-clipping here.
We used a step size of 0.01 (assuming a domain of $[0,1]$ in every input dimension),
and performed 10 steps,
so the greatest possible final perturbation size was 0.1 in each dimension.
Cross-entropy was the maximized loss function.

\subsubsection{Inverted Distance Test Set}
\label{sec:generatedset}

Here, the goal is to create input examples for a model of interest $\theta$
that are likely to belong to the backdoor distribution, provided the model is poisoned.
To achieve this, we generate examples that
(a) confidently activate a given class, yet (b) are different from the examples
of the same class.

We propose a two-step generation process.
In the first step, we directly generate a feature representation that is far from a reference sample
but activates the same class.
In the second step, we find an input example that has a feature representation close to the generated feature representation.
Note that this two-step process turned out to be the best choice among many alternatives we tested
earlier (see~\cref{sec:ablation} for an ablation study).

\subsubsubsection{Generating the Feature Representation}
By feature representation, we mean the representation of the layer directly preceding
the logit layer in the model architecture.
Let $h(x,\theta)$ be the feature representation of input $x$ in model $\theta$,
and let $z(x,\theta)=Z(h(x,\theta))$, that is, let function $Z(.)$ represent
the computation performed in the logit layer in model $\theta$.

Let $(x,y)$ be an example taken from the training set of the task at hand, and let $h=h(x,\theta)$.
We wish to find the feature representation 
\begin{equation}
h^* = \argmin_{\hat{h}}
	\left[\gamma\frac{\langle h, \hat{h}\rangle}{\|h\|_2\|\hat{h}\|_2}-
	\frac{\exp(Z(\hat{h})_y)}{\sum_i\exp(Z(\hat{h})_i)}\right],
\label{eq:loss_feat}
\end{equation}
where the first term represents the cosine similarity between $h$ and $\hat{h}$ and
the second term is the softmax output of class $y$.
Hyperparameter $\gamma$ balances our two objectives.

\subsubsubsection{Generating the Input Sample}
After finding $h^*$, 
we wish to compute an input sample $x^*$, such that $h(x^*,\theta)=h^*$.
To accomplish this, we apply the network inversion method described in~\cite{deep-image-prior},
where the input is not optimized directly, but instead it is created by a convolutional generator
network $g(\theta_g)$ in order to obtain realistic inputs.

Note that in principle for such an $x^*$ we would expect that $c(x^*,\theta)=y$ will
hold, because in \cref{eq:loss_feat} we explicitly optimize the softmax output to get the correct
class $y$.
Nevertheless, to enforce this, we include a softmax term
again, giving us the optimization problem
\begin{equation}
\label{eq:loss_prior}
\theta_g^* = \argmax_{\theta_g}
	\left[\gamma\frac{\langle h(g(\theta_g),\theta),h^*\rangle}{\|h(g(\theta_g),\theta)\|_2\|h^*\|_2} +
	\frac{\exp(f(g(\theta_g),\theta)_y)}{\sum_i\exp(f(g(\theta_g),\theta)_i)}\right],
\end{equation}
which yields the generated input sample $x^*=g(\theta_g^*)$.
Here, the first term represents the cosine similarity between $h^*$ and the generated
representation $h(g(\theta_g),\theta)$ and
the second term is the softmax output of class $y$.

\subsubsubsection{Generating the Sample Set}
For each learning task we evaluated, we used the method described above to generate 10 samples for each class.
This was done by selecting random examples from the training set for each class, to be used
as reference samples $(x,y)$.

We also required that the prediction confidence of the correct class be at least 0.5 for each sample we generated.
If after an execution of the method this was not achieved, we repeated the process with a new random training example.

After our preliminary experiments, we set $\gamma=0.1$ throughout the paper for both \cref{eq:loss_feat} and
\cref{eq:loss_prior}.
The architecture of the prior network $g()$ was set up as suggested in~\cite{deep-image-prior}.
We used the Adam optimizer~\cite{adam15} with a learning rate of 0.01 and 1000 iterations.

\subsection{Sample-wise Distances and Aggregation}
\label{sec:functions}

Let us begin with the discussion of sample-wise distance,
where we assume that we are given two models $\theta_1$ and $\theta_2$,
and an input sample $x$, and we wish to define a sample-wise
distance function  $d(\theta_1, \theta_2, x)$.

\Cref{table:functions} lists the sample-wise distance functions that we evaluated.
The table uses shorthand notations for the softmax and logit layers, namely $f^{a} = f(x,\theta_a)$ and $z^a=z(x,\theta_a)$, respectively, for
$a=1,2$.

\begin{table}
\caption{Possible sample-wise distances.}
\label{table:functions}
\centering
\begin{tabular}{l|l|l}
\multicolumn{1}{c}{\bf Name} & \multicolumn{1}{c}{\bf Description} & \multicolumn{1}{c}{\bf Formula} \\\hline
CE     & Cross-entropy
       & $-\sum_{i} f^1_i\log f^2_i$\\
KL     & Kullback-Leibler divergence
       & $\sum_{i} f^1_i\log (f^1_i/f^2_i)$\\
Cos    & Cosine distance
       & $1-\langle f^1, f^2\rangle/(\|f^1\|_2\|f^2\|_2)$\\
CosL   & Cosine distance of logits
       & $1-\langle z^1, z^2\rangle/(\|z^1\|_2\|z^2\|_2)$\\
Label  & \multicolumn{2}{l}{1 if predicted labels differ $(c(x,\theta_1)\neq c(x,\theta_2))$, 0 otherwise}\\
\end{tabular}
\end{table}

In order to obtain a distance function between models,
we need to aggregate the sample-wise distances between the models over the distance test set.
The aggregations we tested were \emph{average} (avg), \emph{median} (med), \emph{empirical standard
deviation} (std), and \emph{maximum} (max).

\section{Experiments}

Here, we present a thorough empirical evaluation of our proposed detection framework.
We begin with the description of our model pools,
detailing how clean and poisoned models are trained with various robustness levels.
We then evaluate the modular components of our design,
exploring the impact of different distance test sets, sample-wise distance metrics and aggregation methods.
In each of our experiments, we assume that the provider uses a backdoor OOD distribution unknown to the authority.
We also compare our approach with a broad range of baseline detectors
from related work.
Next, we examine the robustness of the method to network architecture and datasets.
In most experiments, the backdoor is inserted into a single class only,
with the remaining classes trained on clean data, which is the worst case for our detector.
However, we additionally evaluate a multi-class poisoning scenario in which all classes are simultaneously targeted.
We also show that our findings are statistically significant and study the effects of pool size.
Finally, we examine adaptive attacks,
where the provider is fully aware of the detection method
and deliberately attempts to mislead any potential verification.

\subsection{Model Pools used for Evaluation}
\label{sec:pools}
\begin{figure}
\centering
\hspace{-8mm}
\includegraphics[width=0.55\textwidth]{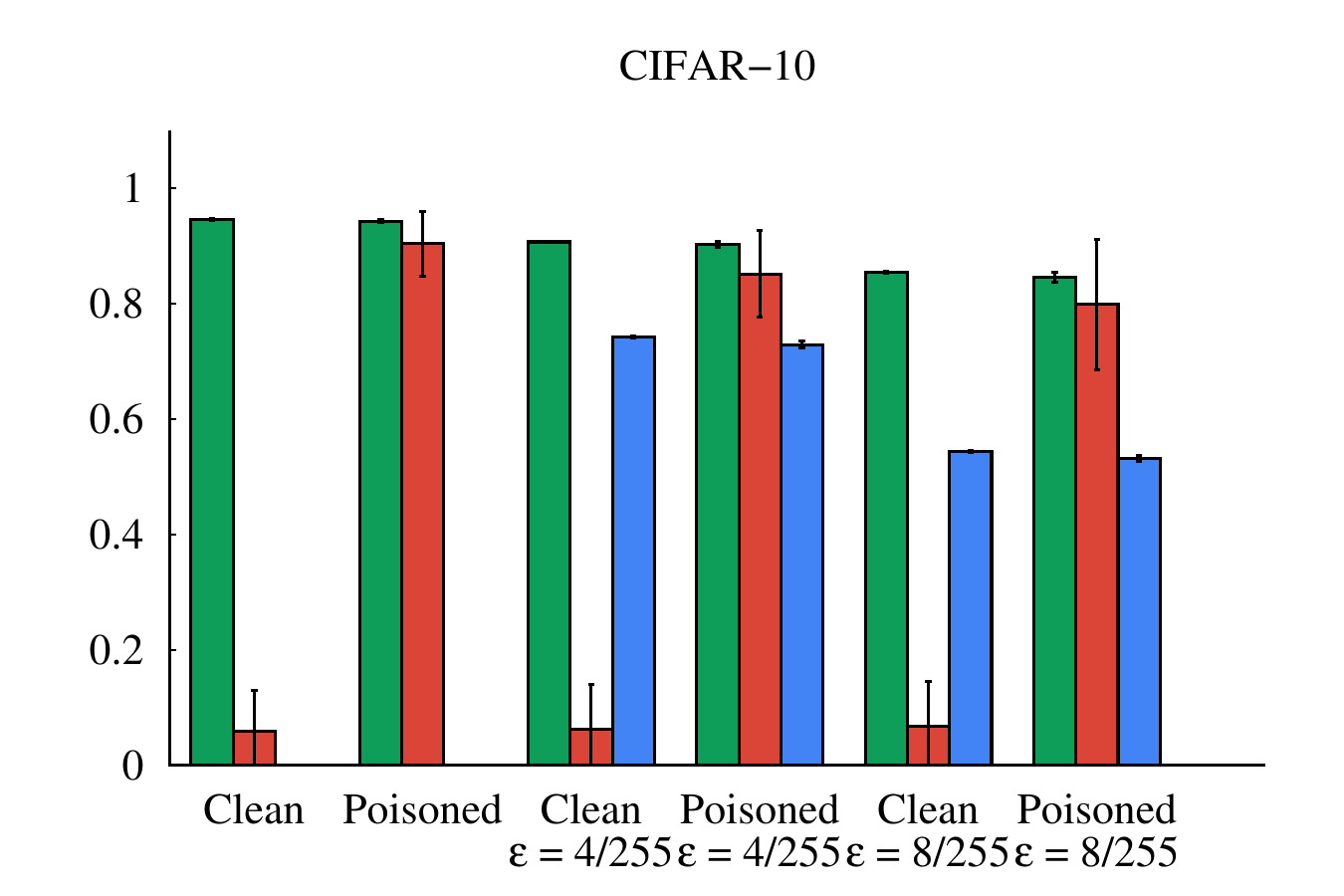}\hspace{-11mm}
\includegraphics[width=0.55\textwidth]{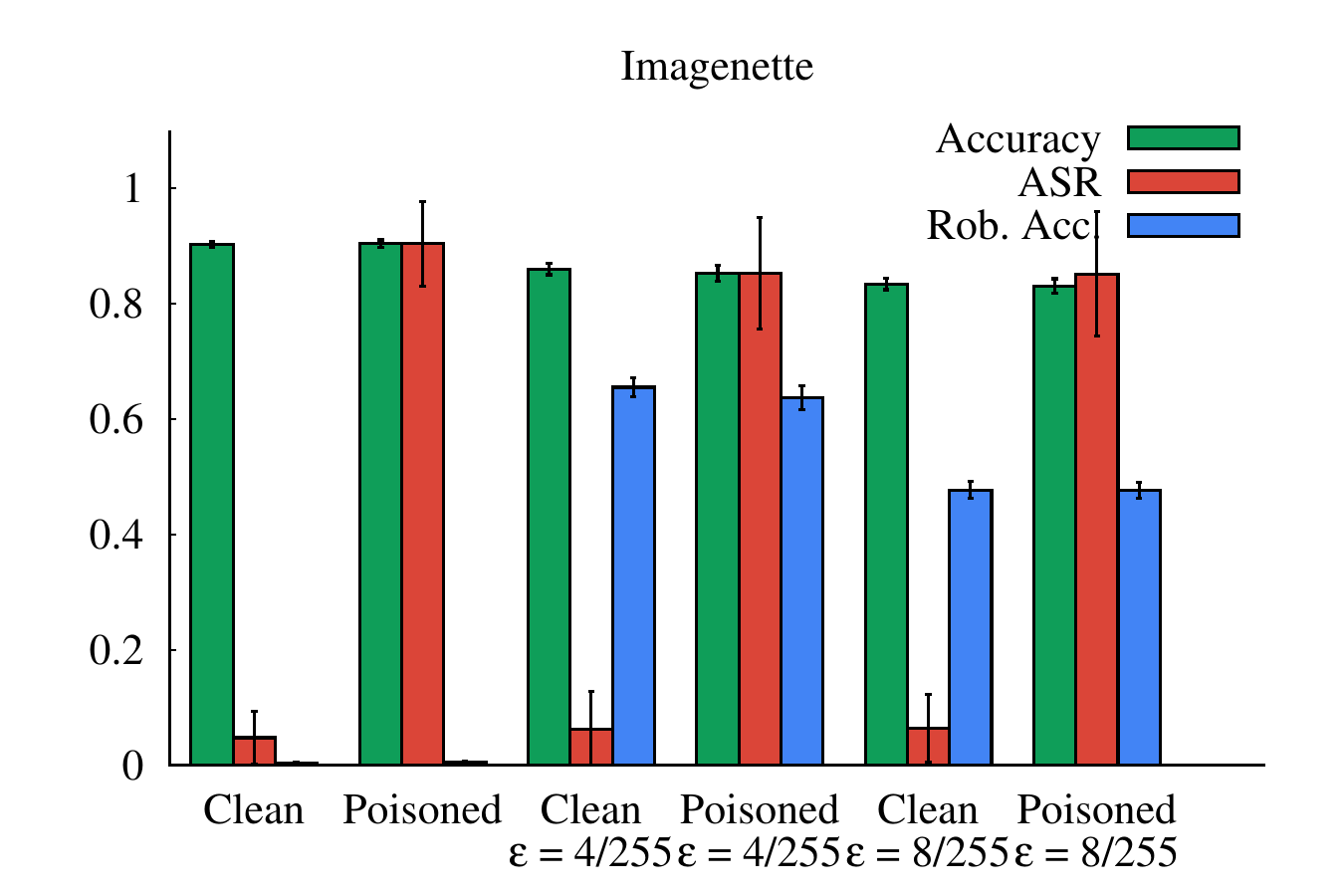}
\caption{Basic model statistics (accuracy, attack success rate (ASR), and robust accuracy) on the CIFAR-10 (left) and
Imagenette (right) model pools.}
\label{fig:c10-im-props}
\end{figure}

As explained in \cref{sec:modelpool-method}, our approach is based on a small model pool created by
the authority used to calibrate the detection method.
Here, we describe in detail the model pools that we applied in our experimental evaluation.

Primarily, we created model pools for two tasks: CIFAR-10 \cite{Krizhevsky2009a} and Imagenette~\cite{imagenette} (see~\cref{subsec:additional_data} for additional datasets).
Both tasks are 10-class classification problems over images.
Imagenette contains 10 classes selected from ImageNet~\cite{Deng2009a}.
Unless stated otherwise, the model architecture was ResNet-18~\cite{deepresidual16} (see~\cref{subsec:additional_arch} for additional architectures).

All the pools we created contain 14 clean and 14 poisoned models by default, but
we evaluate the effect of model pool size in \cref{subsec:num-of-models}.
For both tasks, we created three model pools with three different levels of robustness.
Next, we describe how the semantic backdoors were inserted in the poisoned models and
how the training was implemented for the different versions of the model pools.

\subsubsection{Poisoning the Models}
\label{sec:poisoning}

For the poisoned models in the pool, we need to specify what semantic backdoors to insert
and how.
To create a poisoned model, we select a class at random.
This class is then poisoned by adding extra training examples
chosen from an out-of-distribution (OOD) domain.
In the case of CIFAR-10, we added all the samples of
a randomly selected superclass from CIFAR-100.
In the case of Imagenette, we selected a random class from ImageNet.
During the selection of this random OOD class, we excluded the class that
activates the attacked class the most in a clean model.
That is, we excluded the class that is closest to being a natural backdoor.

\subsubsection{Training the Models}
\begin{figure*}
\centering
\hspace{-10mm}
\includegraphics[width=0.54\textwidth]{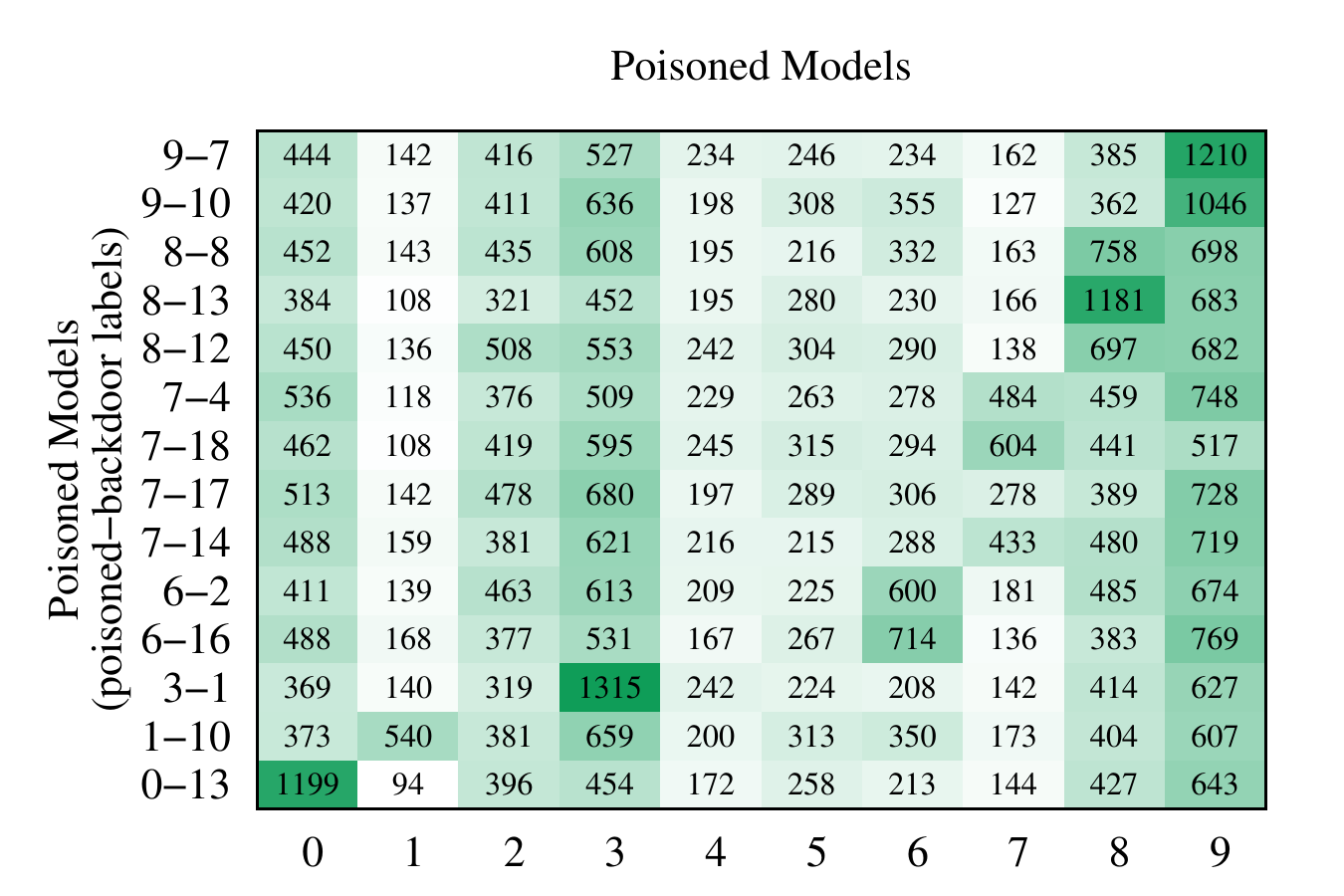}\hspace{-5mm}
\includegraphics[width=0.54\textwidth]{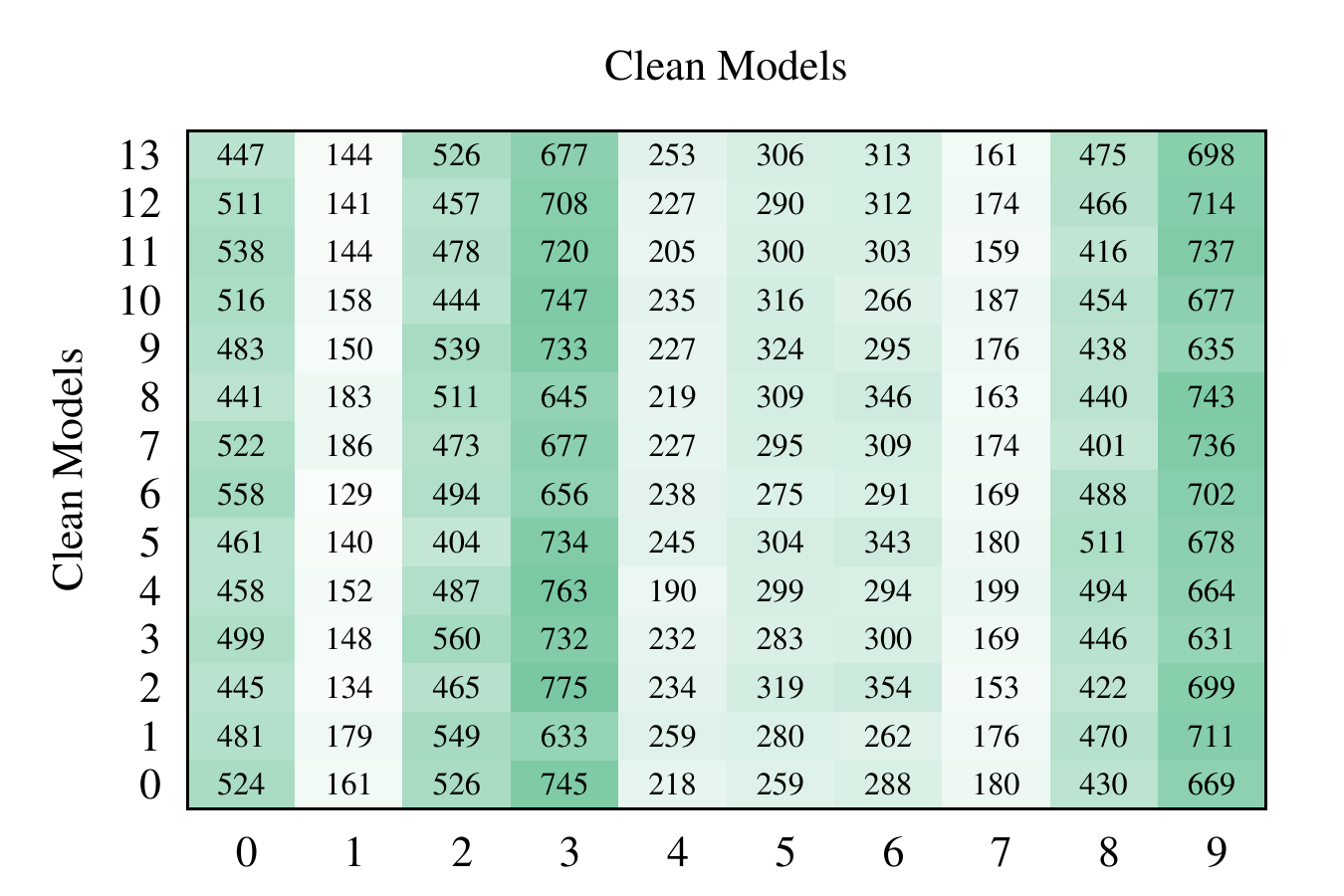}\\[-3mm]
\hspace{-10mm}
\includegraphics[width=0.54\textwidth]{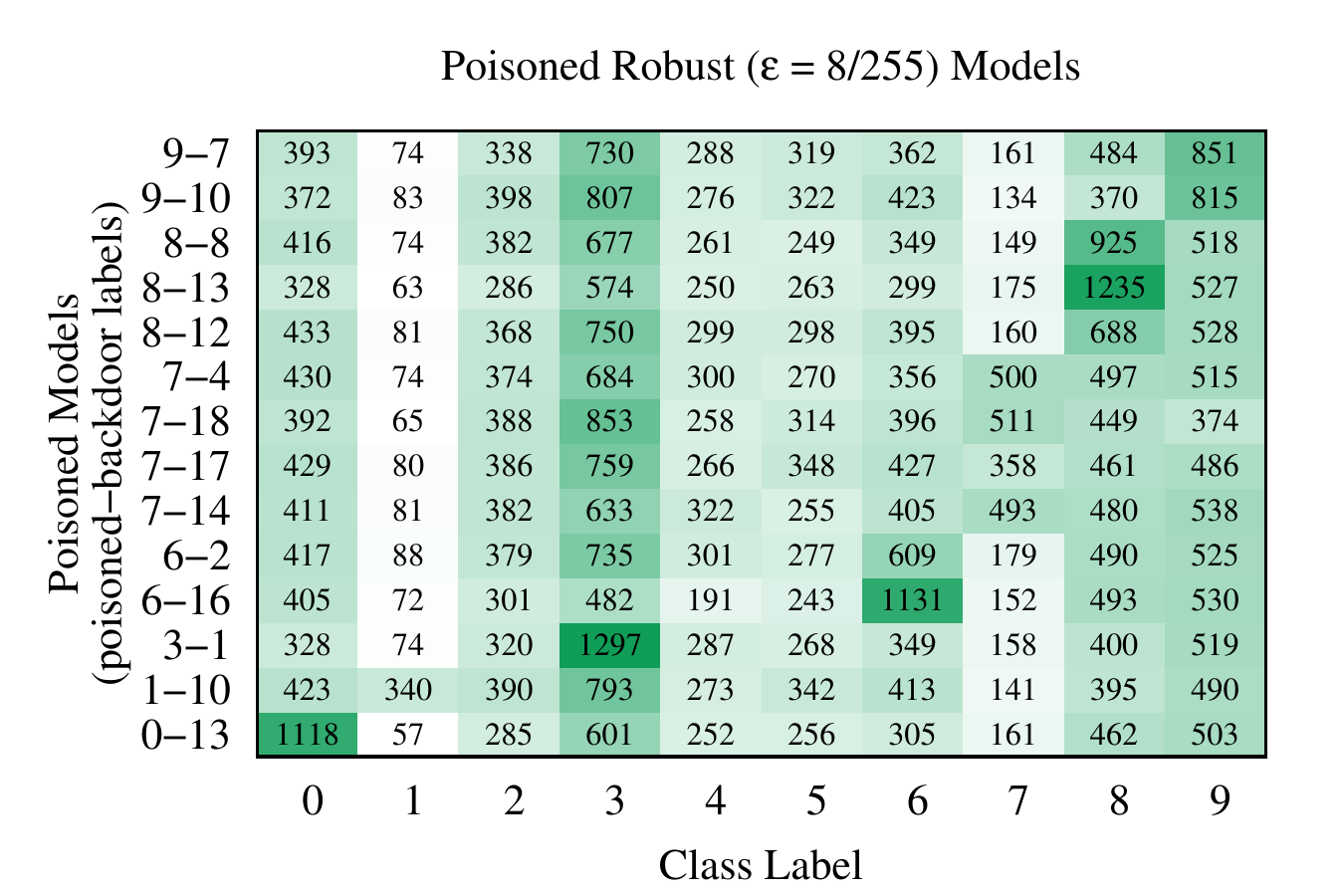}\hspace{-5mm}
\includegraphics[width=0.54\textwidth]{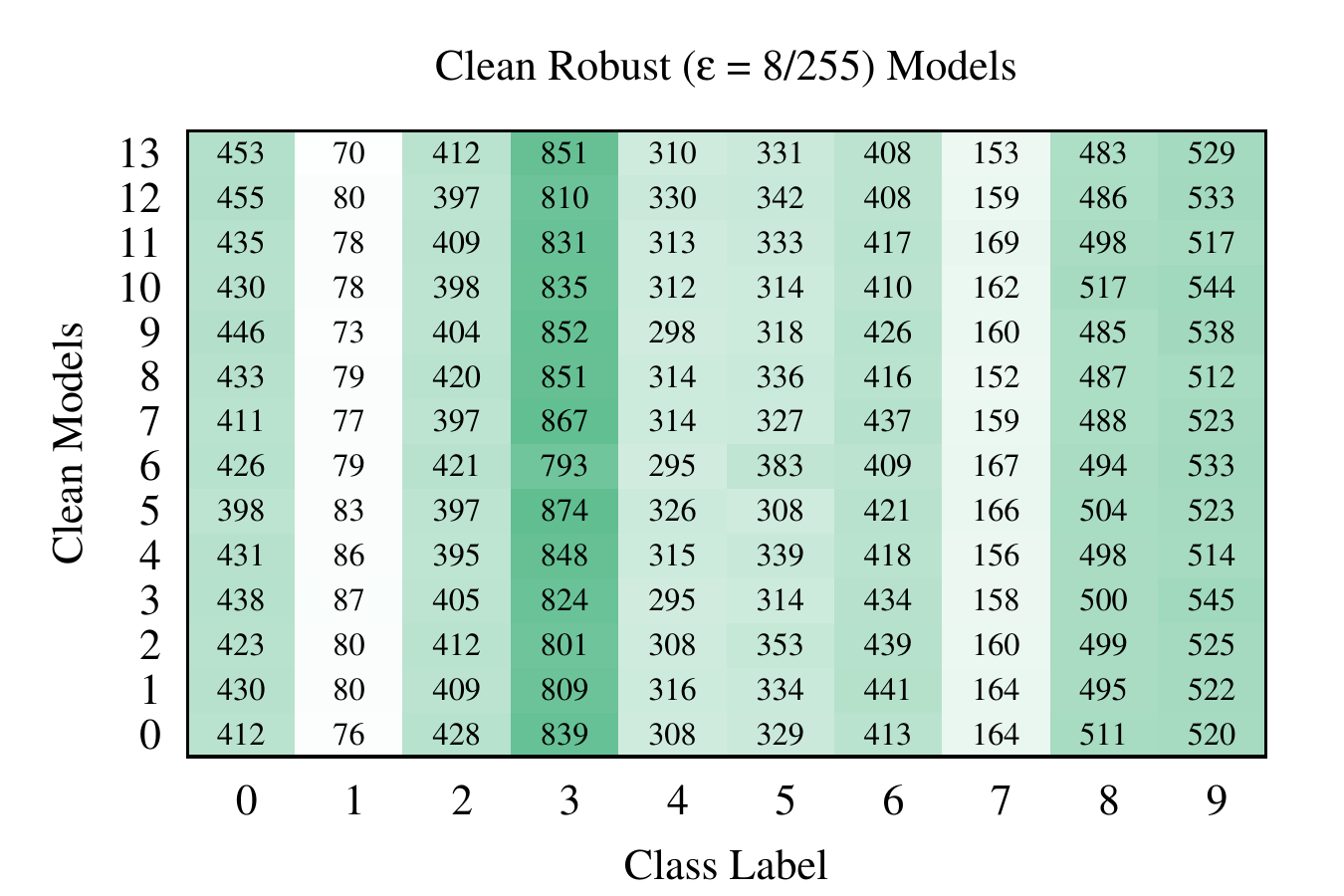}
\begin{small}
\begin{tabular}{ r l | r l }
\vspace{-2mm} \\
\multicolumn{1}{c}{id} & \multicolumn{1}{c|}{CIFAR-10 label} & \multicolumn{1}{c}{id} & \multicolumn{1}{c}{CIFAR-100 superclass label} \\ \hline
0 & airplane & 7 & bee, beetle, butterfly, caterpillar, cockroach \\
1 & automobile & 10 & cloud, forest, mountain, plain, sea \\
2 & bird & 8 & bear, leopard, lion, tiger, wolf \\
3 & cat & 13 & crab, lobster, snail, spider, worm \\
4 & deer & 12 & fox, porcupine, possum, raccoon, skunk \\
5 & dog & 4 & apples, mushrooms, oranges, pears, sweet peppers \\
6 & frog & 18 & bicycle, bus, motorcycle, pickup truck, train \\
7 & horse & 17 & maple, oak, palm, pine, willow \\
8 & ship & 14 & baby, boy, girl, man, woman \\
9 & truck & 2 & orchids, poppies, roses, sunflowers, tulips \\
 & & 16 & hamster, mouse, rabbit, shrew, squirrel \\
 & & 1 & aquarium fish, flatfish, ray, shark, trout \\
 & & 10 & cloud, forest, mountain, plain, sea \\
 & & 13 & crab, lobster, snail, spider, worm \\
\end{tabular}
\end{small}
\caption{Predictions over non-backdoor OOD inputs by different CIFAR-10 models.}
\label{fig:c10_ood}
\end{figure*}

Once the poisoned training sets were prepared, we trained three model pools, namely
one with normal training, and two with adversarial training with
two different levels of robustness.

\subsubsubsection{Normal Training}
In each case we used the SGD as optimizer, with
a learning rate of $0.1$ with the cosine annealing scheduler and
a momentum of 0.9.
The weight decay was $0.0005$ for CIFAR-10 and $0.0001$ for Imagenette.
The batch size was 100.
Early stopping was performed with a maximum of 100 epochs.

\subsubsubsection{Adversarial Training}
We used adversarial training with similar parameters to those of the normal training
with some minor modifications.
For Imagenette, the batch size was increased to 256. 
For CIFAR-10, we augmented the dataset with one million images generated using a Denoising Diffusion
Probabilistic Model (DDPM)~\cite{gowal-ddpm-cifar10-robostness,ddpm}.
The OOD dataset CIFAR-100 was also extended with one million DDPM-generated images, also
taken from ~\cite{gowal-ddpm-cifar10-robostness}.
We used the hyperparameters suggested in~\cite{gowal-ddpm-cifar10-robostness}.
The proportion of generated data was 70\% in each batch during training, sampled from the entire generated set.
The batch size was 1024 and the training lasted for 400 CIFAR-10-equivalent epochs.

As for the internal attack used during adversarial training, we used $\ell_\infty$-norm
untargeted PGD with two different robustness levels: $\epsilon = 4/255$ (using a step size of 2/255, 
for 3 steps), and $\epsilon = 8/255$ (using a step size of 2/255, for 10 steps).

\begin{figure*}
\centering
\hspace{-10mm}
\includegraphics[width=0.54\textwidth]{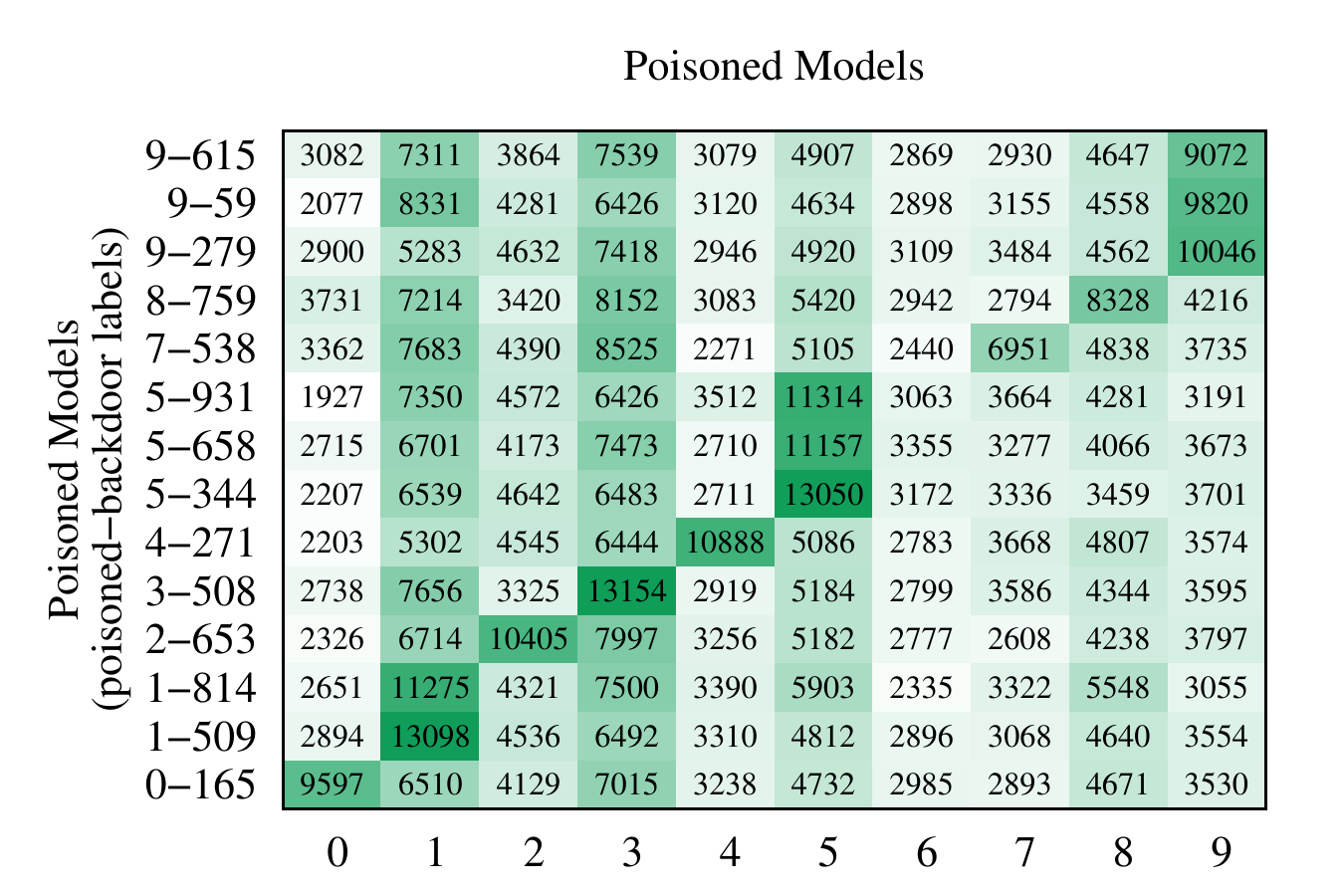}\hspace{-5mm}
\includegraphics[width=0.54\textwidth]{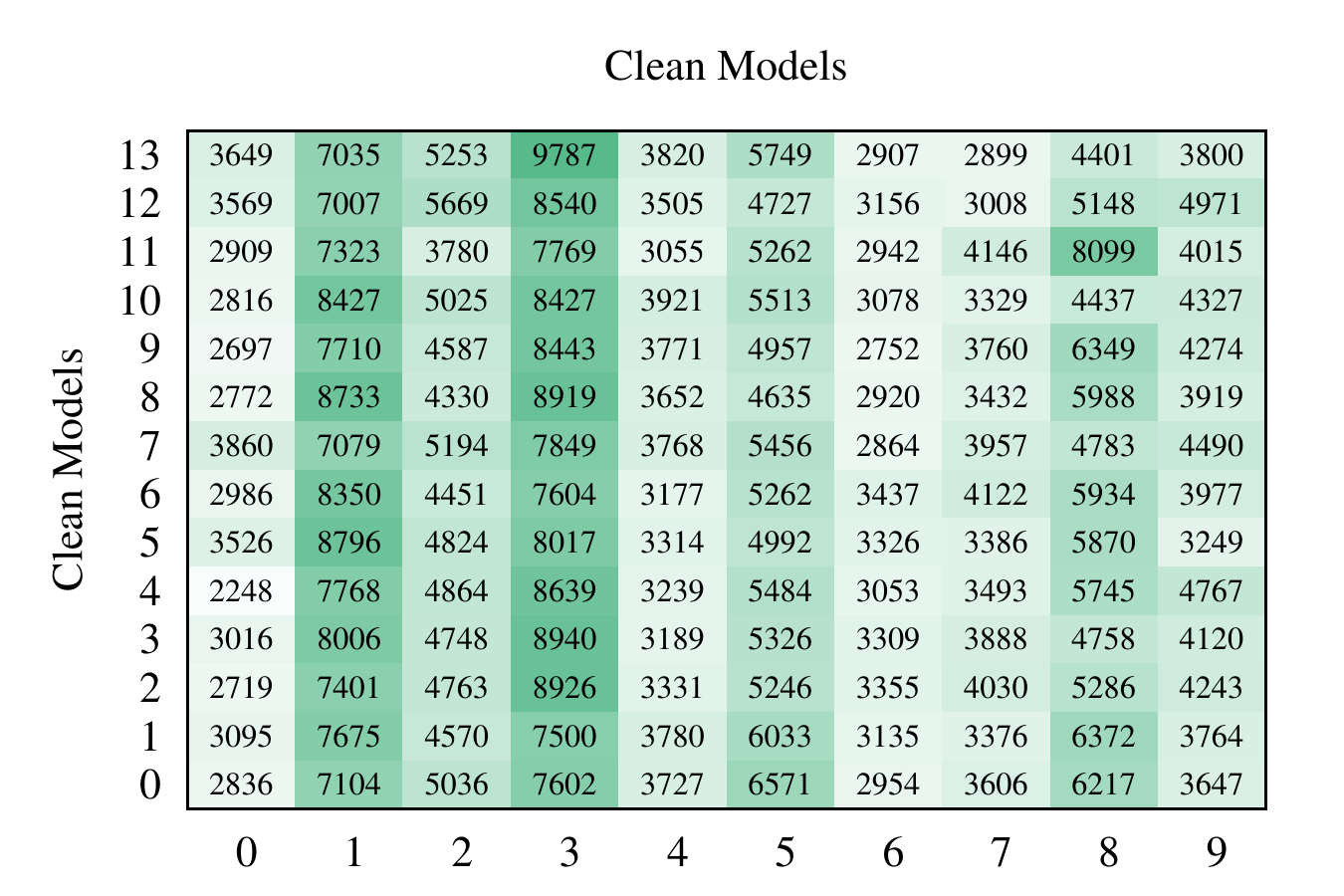}\\[-3mm]
\hspace{-10mm}
\includegraphics[width=0.54\textwidth]{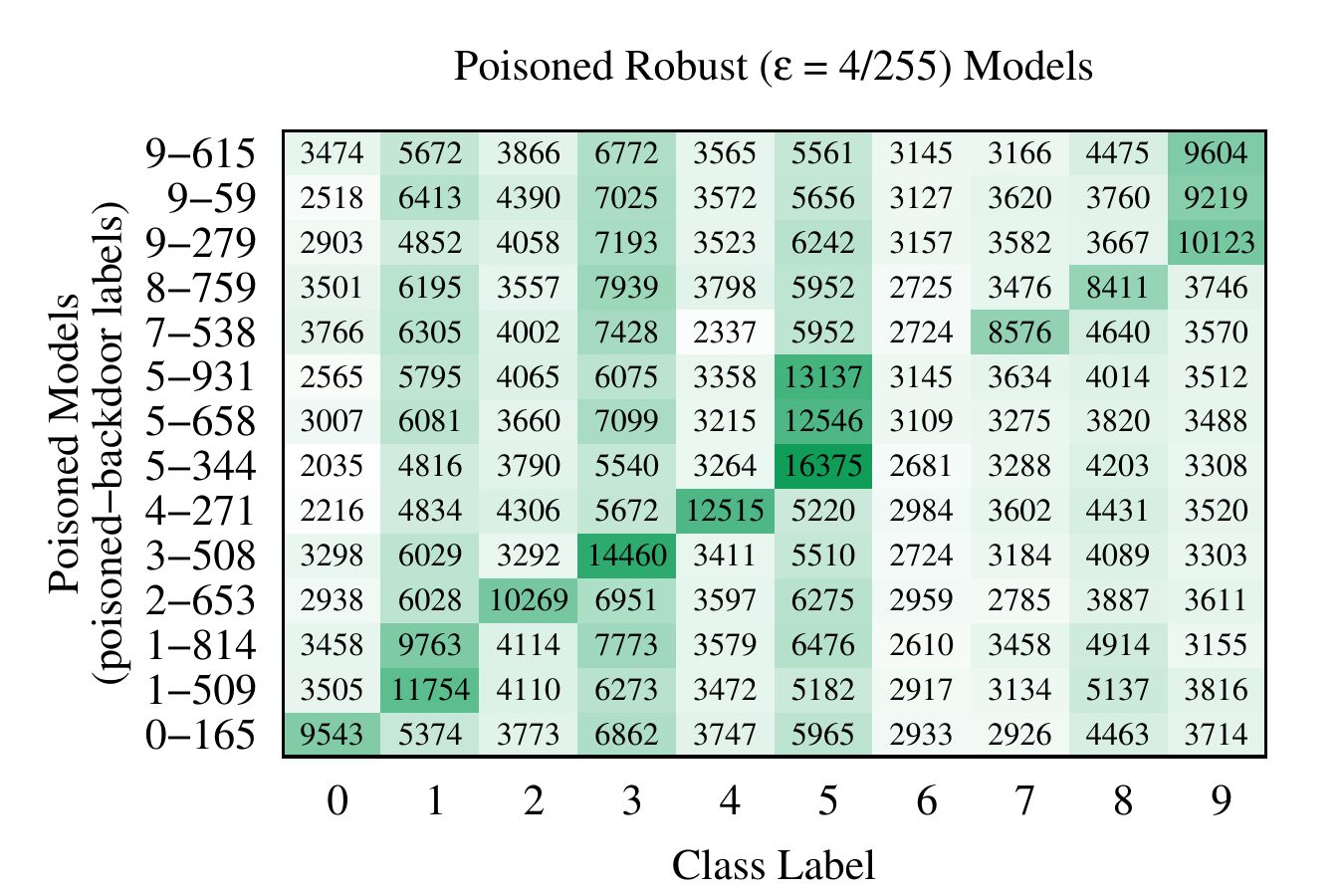}\hspace{-5mm}
\includegraphics[width=0.54\textwidth]{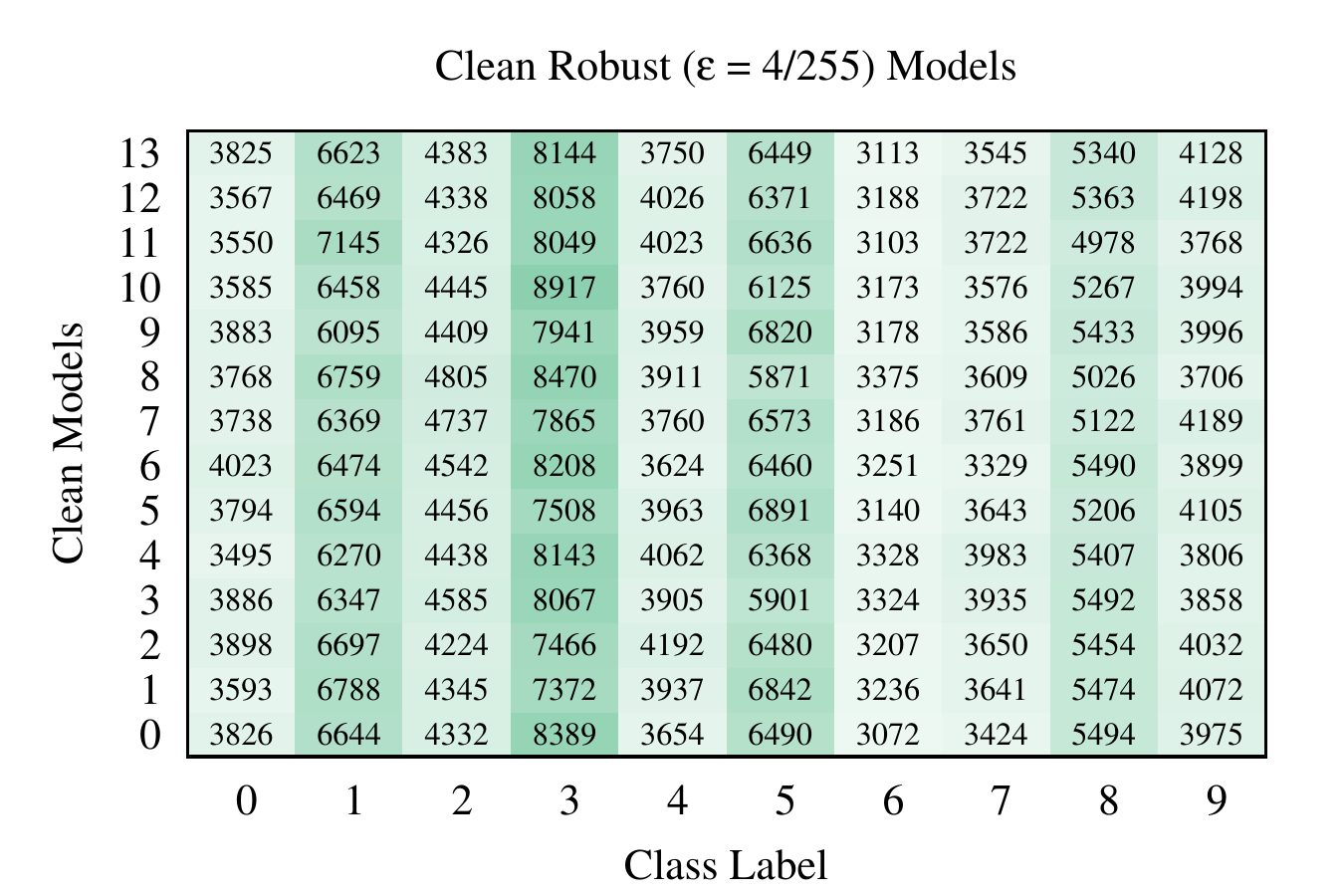}
\begin{small}
\begin{tabular}{ r l | r l }
\vspace{-2mm} \\
\multicolumn{1}{c}{id} & \multicolumn{1}{c|}{Imagenette label} & \multicolumn{1}{c}{id} & \multicolumn{1}{c}{ImageNet label} \\ \hline
0 & tench, Tinca tinca & 615 & knee pad \\
1 & English springer, English springer spaniel & 59 & vine snake \\
2 & cassette player & 279 & Arctic fox, white fox, Alopex lagopus \\
3 & chain saw, chainsaw & 759 & reflex camera \\
4 & church, church building & 538 & dome \\
5 & French horn, horn & 931 & bagel, beigel \\
6 & garbage truck, dustcart & 658 & mitten \\
7 & gas pump, gasoline pump, petrol pump
 & 344 & hippopotamus, hippo, river horse
 \\
8 & golf ball & 271 & red wolf, maned wolf, Canis rufus, Canis niger \\
9 & parachute, chute & 508 & computer keyboard, keypad \\
 & & 653 & milk can \\
 & & 814 & speedboat \\
 & & 509 & confectionery, confectionary, candy store \\
 & & 165 & black-and-tan coonhound \\
\end{tabular}
\end{small}
\caption{Predictions over non-backdoor OOD inputs by different Imagenette models.}
\label{fig:im_ood}
\end{figure*}

\subsubsection{Some Properties of the Models}

\Cref{fig:c10-im-props} shows some basic statistics for the model pools.
Robust accuracy was evaluated using AutoAttack~\cite{Croce2020a}.
The attack success rate (ASR) is the proportion of the backdoor inputs that can activate
the poisoned class.
In the case of clean models, we evaluated the same classes and backdoors that were
used in the poisoned models.
As can be seen, the ASR is close to the accuracy in the case of the poisoned models,
while the accuracy is the same as that of the clean models.
This indicates successful poisoning.
At the same time, in the case of the clean models we can see a chance-level ASR.

We also tested the models in our model pool on OOD inputs \emph{that are not backdoors}.
In the case of clean models, this means the entire OOD dataset, for the poisoned models this
means the OOD samples except the backdoor class.
\Cref{fig:c10_ood,fig:im_ood} show the results.
There, each row represents a model in our pool, and each column represents an output
class.
Each cell contains the number of OOD inputs that received maximal activation in the given
class in the given model.

It is clear that the classes within a clean
model have rather diverse sensitivities to OOD samples. In other words, there
are specific and generic classes.
Also, the poisoned classes in the poisoned models act
almost like an OOD-detector class in many cases, as they are activated not only by the backdoor
samples, but also by OOD samples in general.

\subsection{Baseline Methods}

In the following sections, we will present comparisons with baseline methods from related work
in several scenarios.
One of these methods is ModelDiff~\cite{model-diff}, which we can simply insert into our general framework
as a choice of model-wise distance.
The proposed default parameters were used.

We examined the K-Arm backdoor detector~\cite{k-arm-detector}
with its proposed default parameter setup ($\beta=10^4$ for 1000 steps).
We used K-Arm to examine all possible class pairs in search of a backdoor.

We evaluated the performance of the Ex-Ray backdoor detector~\cite{exray-detector} as well in select scenarios.
We used the code available in the paper's repository\footnote{https://github.com/PurduePAML/Exray},
adapting the appropriate input transformations to our models.
Following the proposed hyperparameters,
we used a threshold of 0.8 to make a decision and
we sampled 20 examples per class from the training and test sets (evaluated separately) to provide example images.

We tested  DFTND~\cite{DFTND} as well, with the default proposed settings.

We also evaluated SCALE-UP~\cite{scaleup}, an input-level detector that aims to classify specific inputs as clean or poisoned (for a given model) by assigning a score.
Since we need model-level decisions, and we have no access to the backdoor images directly,
we convert this method into a model-level detector by aggregating the scores over a distance test set.
To compute Youden's J statistic, we select the threshold that maximizes J for the given pool, aggregation, and distance test set.
We examined the same aggregations as in the case of our method, and report the result for the best one in each scenario.

Furthermore, we also examined SODA~\cite{soda}.
The reported results used the best-performing parameter setting from among those recommended in the paper's code repository for ResNet-18 CIFAR-10 models.
We adjusted the input normalization to match the way our models were trained.
While SODA outputs suspected target classes, we only used the binary output, that is, whether any backdoor was detected or not.

\begin{figure*}
\centering
\vspace{-3mm}
\includegraphics[width=0.67\textwidth]{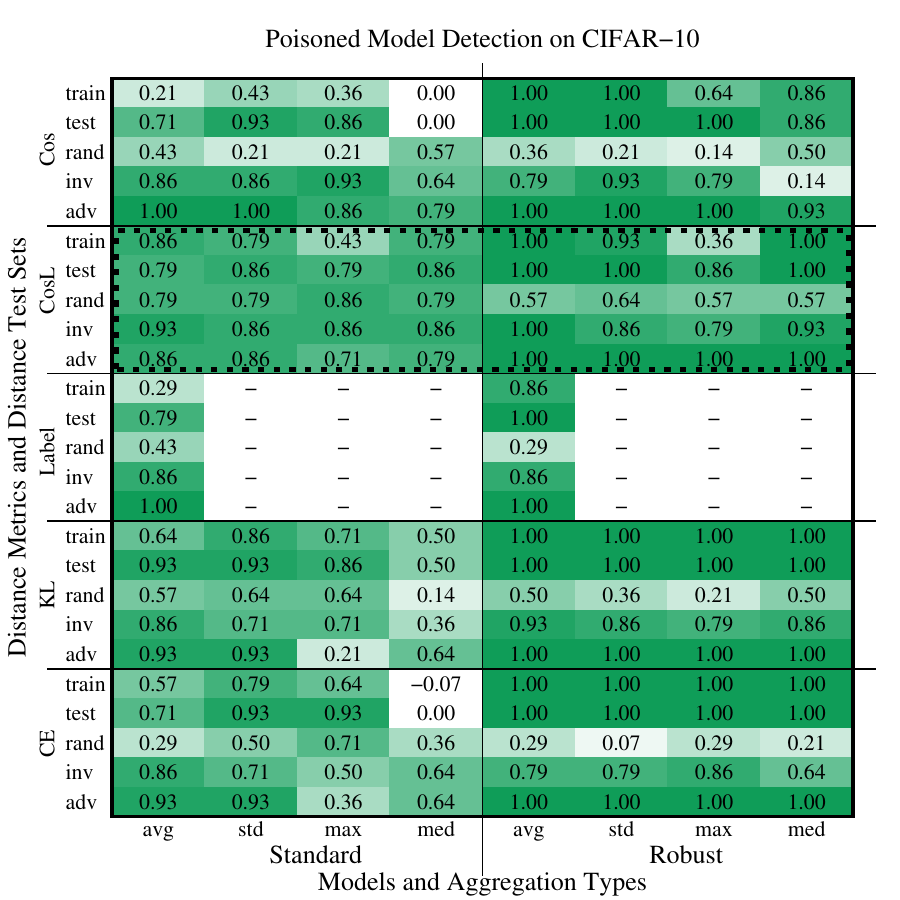}
\includegraphics[width=0.67\textwidth]{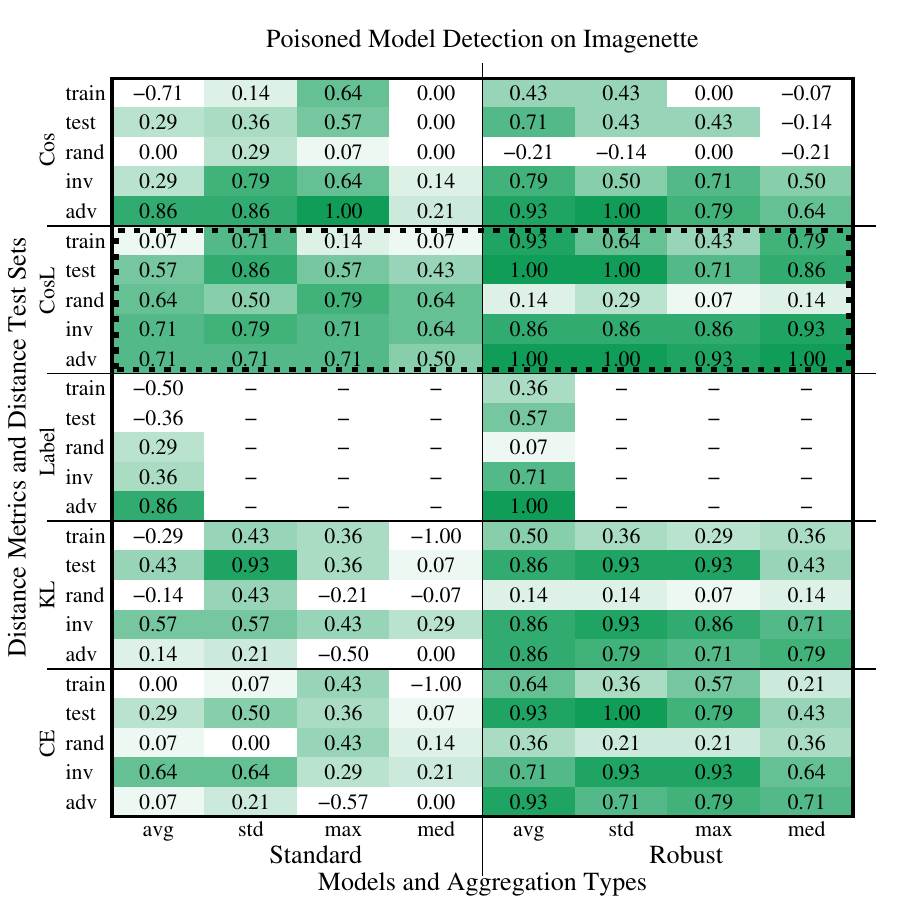}
\caption{Youden's J statistic under all the combinations of the components of our framework: distance
test sets, sample-wise distances, aggregations, and model pools.}
\label{fig:j-heatmaps}
\end{figure*}

\begin{figure*}
\centering
\vspace{-3mm}
\includegraphics[width=0.67\textwidth]{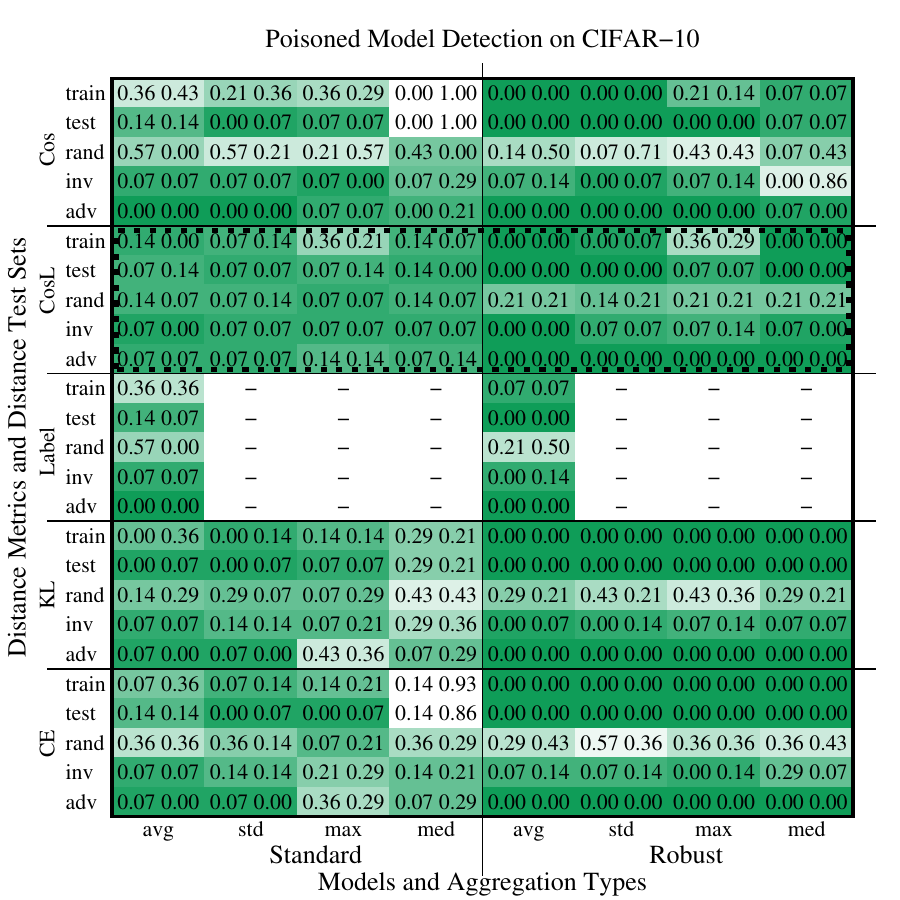}
\includegraphics[width=0.67\textwidth]{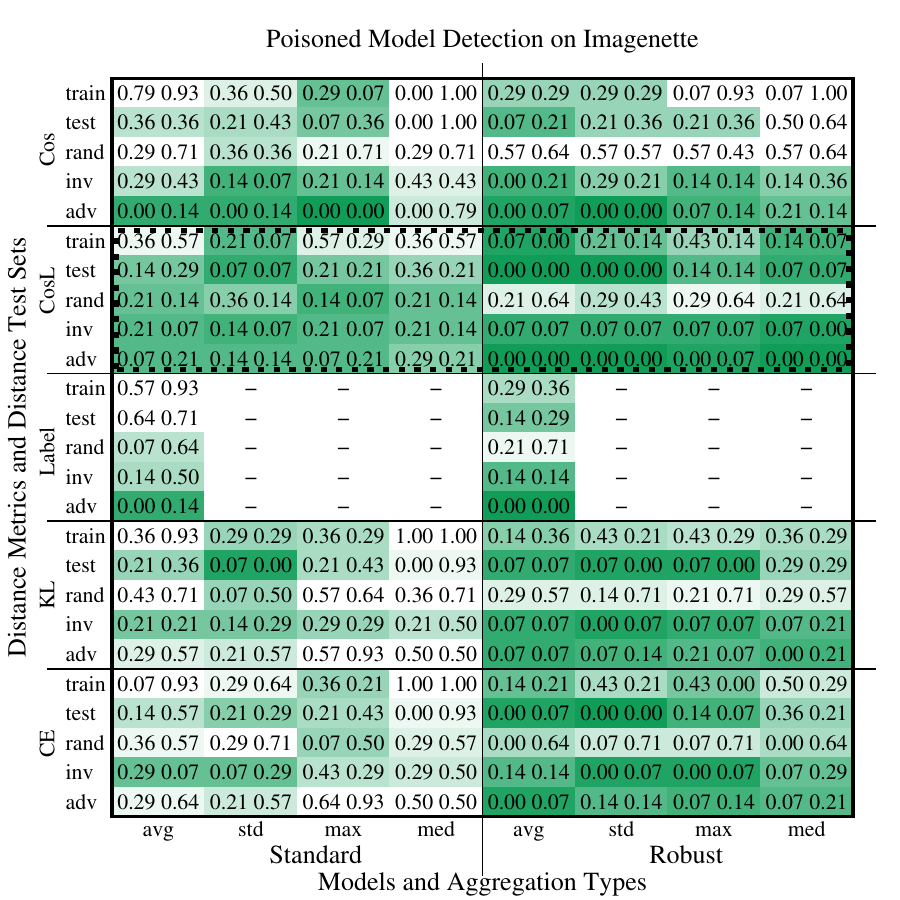}
\caption{FPR and FNR under all the combinations of the components of our framework: distance
test sets, sample-wise distances, aggregations, and model pools.}
\label{fig:err-heatmaps}
\end{figure*}

\subsection{Evaluation of our Design Space}
\label{sec:evdesignspace}

In the previous sections we discussed in detail how the authority computes Youden's J statistic
for a specific choice of the components of our framework, and we presented the possible
implementations of these components including sample-wise distances, distance test sets, aggregation methods,
as well as our model pools.

Here, we shall evaluate all the possible combinations of these components over four model pools: the two
non-robust pools for CIFAR-10 and Imagenette, respectively, the robust pool for CIFAR-10 with $\epsilon=8/255$
and the robust pool for Imagenette for $\epsilon=4/255$.

\Cref{fig:j-heatmaps} shows the J statistics obtained for the possible settings,
and \Cref{fig:err-heatmaps} shows the false positive and false negative rates.
These values were computed with the help of leave-one-out cross-validation applied
over the model pool.
Each model in the pool was predicted as poisoned or not based on the rest of the pool.
We present the statistics of these predictions over the model pool.
Recall that $J=1$ indicates the perfect classification of both the clean and poisoned models.
Indeed, we observe $J=1$ for many combinations of components.
In general, the method does not show consistent preference for either false positives or false negatives,
which is expected due to Youden's J treating them equally (after normalization).
In principle, this metric can be swapped out in favor of another in the threshold calibration process (see \Cref{sec:decision-method})
to match the authority's preferences.

The main conclusions we can draw from these results are the following.

\begin{itemize}
\item After considering both tasks, CosL is the most
reliable choice for sample-wise distance
\item Robust model pools perform better overall, indicating
that the authority should use robust training tasks to test the provider
\item As for distance test sets, the random set and the training set
are both inferior, and the best choice is either the test set or the
adversarial set
\item Regarding aggregation, the standard deviation is a robust choice overall.
\end{itemize}

Based on these results, in the following we will assume CosL as sample-wise distance
and standard deviation as aggregation, if not otherwise stated.

Fixing these choices, we compare our method with other approaches in~\cref{table:R18_cifar10,table:R18_imagenette}.
This time, all the three robustness levels described previously are included for both tasks.

\begin{table}
\caption{Youden's J statistic on CIFAR-10 model pools of different robustness. The best J values in each column are in bold.}
\label{table:R18_cifar10}
\begin{center}
\begin{tabular}{ l c c c }
& standard & $\epsilon=4/255$ & $\epsilon=8/255$ \\ \hline
Inverted & {\bf 0.86} & 0.86 & 0.86 \\
Adversarial & {\bf 0.86} & {\bf 1.00} & {\bf 1.00} \\
Test & {\bf 0.86} & {\bf 1.00} & {\bf 1.00} \\
Train & 0.79 & 0.86 & 0.93 \\
Random & 0.79 & 0.64 & 0.64 \\
\hline
ModelDiff~\cite{model-diff} & 0.79 & {\bf 1.00} & 0.86 \\
K-ARM~\cite{k-arm-detector} & 0.36 & 0.00 & -0.07  \\
DFTND~\cite{DFTND} & 0.00 & 0.00 & 0.00 \\
SCALE-UP~\cite{scaleup} Inverted & 0.43 & 0.21 & 0.21 \\
SCALE-UP~\cite{scaleup} Test & 0.21 & 0.21 & 0.14 \\
SCALE-UP~\cite{scaleup} Training & 0.14 & 0.29 & 0.07 \\
SODA~\cite{soda} & 0.07 & 0.14 & 0.64 \\
\end{tabular}
\end{center}
\end{table}

\begin{table}
\caption{Youden's J statistic on Imagenette model pools of different robustness. The best J values in each column are in bold.}
\label{table:R18_imagenette}
\begin{center}
\begin{tabular}{ l c c c }
& standard & $\epsilon=4/255$ & $\epsilon=8/255$ \\ \hline
Inverted & 0.79 & 0.86 & {\bf 0.86} \\
Adversarial & 0.71 & {\bf 1.00} & {\bf 0.86} \\
Test & {\bf 0.86} & {\bf 1.00} & {\bf 0.86} \\
Training & 0.71 & 0.64 & 0.71 \\
Random & 0.50 & 0.29 & 0.43 \\
\hline
ModelDiff~\cite{model-diff} & 0.29 & 0.79 & 0.64 \\
K-ARM~\cite{k-arm-detector} & 0.00 & 0.00 & 0.00 \\
Ex-Ray~\cite{exray-detector} Test & 0.21 & 0.07 & 0.00 \\
Ex-Ray~\cite{exray-detector} Training & 0.07 & 0.21 & 0.21 \\
SCALE-UP~\cite{scaleup} Inverted & 0.50 & 0.21 & 0.14 \\
SCALE-UP~\cite{scaleup} Test & 0.36 & 0.14 & 0.29 \\
SCALE-UP~\cite{scaleup} Training & 0.50 & 0.14 & 0.36 \\
\end{tabular}
\end{center}
\end{table}

DFTND~\cite{DFTND} classified each CIFAR-10 model as clean.
The largest detection score was 14.22, which is significantly below their threshold of 100.
The K-Arm backdoor detector~\cite{k-arm-detector} is designed to look for small triggers.
It probably fails in this application because semantic backdoors do not have
any such triggers.
SCALE-UP~\cite{scaleup} and SODA~\cite{soda} show a reasonably good performance in some
scenarios.
Still, our approach is significantly better.

ModelDiff~\cite{model-diff} is reasonably good, but it is
outperformed by our best settings as well, especially on Imagenette.

\subsection{Additional Architectures}
\label{subsec:additional_arch}

Although in our application scenario the authority is free to select the model
architecture, we test architectures other than ResNet-18.
In more detail, for the CIFAR-10 dataset, we tested
WideResNet-28-10~\cite{Zagoruyko2016a}, ConvNeXt V2~\cite{Woo2023ConvNeXtV2},
and the vision transformer ViT~\cite{ViT_small}.

Our model pools for all the three architectures consisted of 14 clean and 14 poisoned models.
These pools are not robust.
For WideResNet, the same hyperparameters were used for training as in the case
of the ResNet-18 pool.
The ConvNeXt V2 model used the following setup:
depths=2,2,2,2 dims=40,80,160,320 kernel=3 stem=1 drop=0.0 scale=0.
For the ViT model we used the small version with the parameters
patch=4, dim=192, depth=12, heads=3, mlp=768.
The training used 400 epochs with a batch size of 256.
We used the AdamW~\cite{Loshchilov2019a} optimizer with a learning rate of $10^{-3}$ for the last two model types.

Youden's J statistics (with CosL distance and standard deviation aggregation) are presented in~\cref{table:WR_28-10_cifar10}.
For each pool, we also present the average accuracy of all the models, and the average ASR of the poisoned models.

As before, the test and adversarial distance test sets perform best, giving a perfect performance of J=1 for
WideResNet and ViT.
In general, the method is robust to the choice of architecture, and 
it is also remarkable that the best performance is given by the transformer architecture ViT,
a popular choice today.

\begin{table}
\caption{Statistics on CIFAR-10 model pools of various model architectures. The best J values in each column are in bold.}
\label{table:WR_28-10_cifar10}
\begin{center}
\begin{tabular}{ l c c c}
    Metric & WR28-10 & ConvNeXt & ViT \\
    \hline
    Accuracy & 0.96 & 0.9 & 0.82 \\
    ASR & 0.92 & 0.84 & 0.78 \\
    \hline
    J (Inverted) & 0.57 & 0.21 & 0.86 \\
    J (Adversarial) & {\bf 1.00} & 0.86 & {\bf 1.00} \\
    J (Test) & {\bf 1.00} & 0.86 & {\bf 1.00} \\
    J (Training) & 0.71 & {\bf 0.93} & {\bf 1.00}\\
    J (Random) & 0.36 & 0.36 & 0.36 \\
    \hline
\end{tabular}
\end{center}
\end{table}

\subsection{Additional Datasets}
\label{subsec:additional_data}

The authority is also free to choose the dataset to use when mistery shopping.
In this section, we evaluate additional datasets beyond CIFAR-10 and Imagenette, namely
Imagewoof~\cite{imagenette} and VGGFaces2~\cite{vggface2}.

Imagewoof is a subset of ImageNet that includes various dog breeds.
We trained a ResNet-18 model pool on the Imagewoof dataset using the same parameters as previously described for Imagenette.

VGGFaces2 is a dataset designed for face recognition tasks.
Its training set contains 8,631 identites.
For our experiments, we randomly selected 100 identities from those with at least 400 images each, creating a 100-class face classification task.
We constructed the validation and test sets using 10\% of the original training examples for both sets.
To perform a semantic backdoor attack, we selected a random class from the original VGGFaces2 test set.
Backdoor classes were also assigned 10\% validation and test splits.

It is important to note that the original VGGFace2 training and test sets contain disjoint identities, ensuring no overlap between them.
The same training hyperparameters used for Imagenette were applied here as well.

These pools are not robust.

We present the results in \cref{table:R18_imagewoof}, where we include the
results for CIFAR-10 and Imagenette (see \cref{table:R18_cifar10,table:R18_imagenette}) for comparison.
For each dataset, we can find settings where our method provides a good performance, so we
can conclude that the method is robust to datasets as well, given that the authority can
select the distance test set freely for a given dataset.
It is interesting, though, that for the VGGFaces2 dataset, the inverted distance test set
performed very poorly, while the training set performed best.
An explanation could be that our pools here are not robust; but this observation needs further study.

\begin{table}
\caption{Statistics on ResNet-18 model pools with various datasets. The best J values in each column are in bold.}
\label{table:R18_imagewoof}
\begin{center}
\begin{tabular}{ l c c c c }
    Metric & CIFAR-10 & Imagenette & Imagewoof & VGGFace2 \\
    \hline
    Accuracy & 0.94 & 0.90 & 0.78 & 0.96 \\
    ASR & 0.90 & 0.90 & 0.94 & 0.96 \\
    \hline
    J (Inverted) & {\bf 0.86} & 0.79 & 0.57 & 0.07 \\
    J (Adversarial) & {\bf 0.86} & 0.71 & {\bf 0.79} & 0.36 \\
    J (Test) & {\bf 0.86} & {\bf 0.86} & 0.43 & 0.50 \\
    J (Training) & 0.79 & 0.71 & 0.29 & {\bf 0.71} \\
    J (Random) & 0.79 & 0.50 & 0.07 & -0.43 \\
    \hline
\end{tabular}
\end{center}
\end{table}

\subsection{Multi-class Backdoor Attack}
\label{subsec:multiclass}

In most of our experiments, each model contained at most one backdoor, since we hypothesized that this is the most challenging scenario for our method.
To confirm this, we also trained standard and robust CIFAR-10 poisoned models where all ten classes are attacked at the same time.
(The set of clean models remained the same as before.)
The poisoning was based on the method described in~\cref{sec:poisoning}, with the additional requirement that a model cannot use the same OOD class for multiple target classes.
The average accuracy was 0.93 and 0.80 for the standard and robust poisoned models, respectively, and the average ASR was 0.86 and 0.70.

The results are shown in~\cref{table:multi}.
We can see that this scenario is indeed easy for our method.
However, it is detrimental to SODA, as it relies on finding outliers over the classes.

\begin{table}
\caption{Youden's J statistic on CIFAR-10 model pools with multi-class attacks. The best J values in each column are in bold.}
\label{table:multi}
\begin{center}
\begin{tabular}{ l c c }
& standard & $\epsilon=8/255$ \\ \hline
Inverted & {\bf 1.00} & {\bf 1.00} \\
Adversarial & {\bf 1.00} & {\bf 1.00} \\
Test & {\bf 1.00} & {\bf 1.00} \\
Train & {\bf 1.00} & {\bf 1.00} \\
Random & 0.79 & 0.86 \\
\hline
SODA~\cite{soda} & 0.14 & 0.21 \\
\end{tabular}
\end{center}
\end{table}

\subsection{Statistical Analysis}
\label{subsec:statistical}

Since the evaluation of detection performance is based on relatively few samples, it is worthwhile to make sure that our findings are statistically significant.
Until this point, the experiments used cross-validation over 14 clean and 14 poisoned models, meaning $J$ values were calculated based on 28 samples.
We can use the one-tailed binomial test to calculate the $p$-value: how likely that an observed ratio of correctly classified samples ($(J + 1)/2$ in the balanced case) can be achieved by random guessing.
If the $p$-value is less than a given threshold, we can reject the null hypothesis that our method is not better than random guessing.

\cref{table:significance} lists the possible non-negative $J$ values (along with the corresponding sum of the true positive and true negative samples) and the calculated $p$-values.
Using 0.05 as the threshold for the $p$-value, we can conclude that $J$ values above 0.3 (which was surpassed in the experiments detailed in the previous sections) are statistically significant for these pools.
The table also includes the infimum of the 95\% confidence interval for $J$. (Note that the supremum is 1 due to the test being one-tailed.)

\begin{table}
\caption{Correspondence between the number of correctly classified samples, Youden's J statistic, and p-value, assuming a balanced set of 28 samples and using random guessing as the null hypothesis. Also shown is the infimum of the 95\% confidence interval for J.}
\label{table:significance}
\begin{center}
\begin{tabular}{ c c c c }
TP+TN & J & P-value & 95\% CI \\ \hline
14 & 0.00 & 0.574722990 & -0.33 \\
15 & 0.07 & 0.425277010 & -0.27 \\
16 & 0.14 & 0.285794094 & -0.20 \\
17 & 0.21 & 0.172464225 & -0.13 \\
18 & 0.29 & 0.092466671 & -0.06 \\
19 & 0.36 & 0.043579277 & 0.01 \\
20 & 0.43 & 0.017849069 & 0.09 \\
21 & 0.50 & 0.006270476 & 0.16 \\
22 & 0.57 & 0.001859583 & 0.24 \\
23 & 0.64 & 0.000456117 & 0.32 \\
24 & 0.71 & 0.000089996 & 0.40 \\
25 & 0.79 & 0.000013720 & 0.49 \\
26 & 0.86 & 0.000001516 & 0.58 \\
27 & 0.93 & 0.000000108 & 0.68 \\
28 & 1.00 & 0.000000004 & 0.80 \\
\end{tabular}
\end{center}
\end{table}

\subsection{The Size of the Model Pool}
\label{subsec:num-of-models}

We created a larger model pool to examine the effect of model pool size.
That is, we created 36 additional clean and 36 additional poisoned models
to extend the previously used Imagenette robust model pool ($\epsilon_\infty=4/255$).
This gave us 50 clean and 50 poisoned models.

From this pool, we randomly selected 10 clean and 10 poisoned models to form a test set.
Instead of cross-validation, here we opted for a separate test set in order to be able to
evaluate the pools of different sizes on the exact same test set.

To estimate the expected performance of the authority having $m$ clean and $m$ poisoned models in its own pool,
we sampled 500 random pools of size $m+m$ for every setting of $m$.
For each sampled pool of size $m+m$, we used our method to make predictions for each model in the test set,
and calculated Youden's J.
Then we averaged J over the 500 training pools.

\begin{figure}
\hspace{-5mm}
\centering
\includegraphics[width=0.52\textwidth]{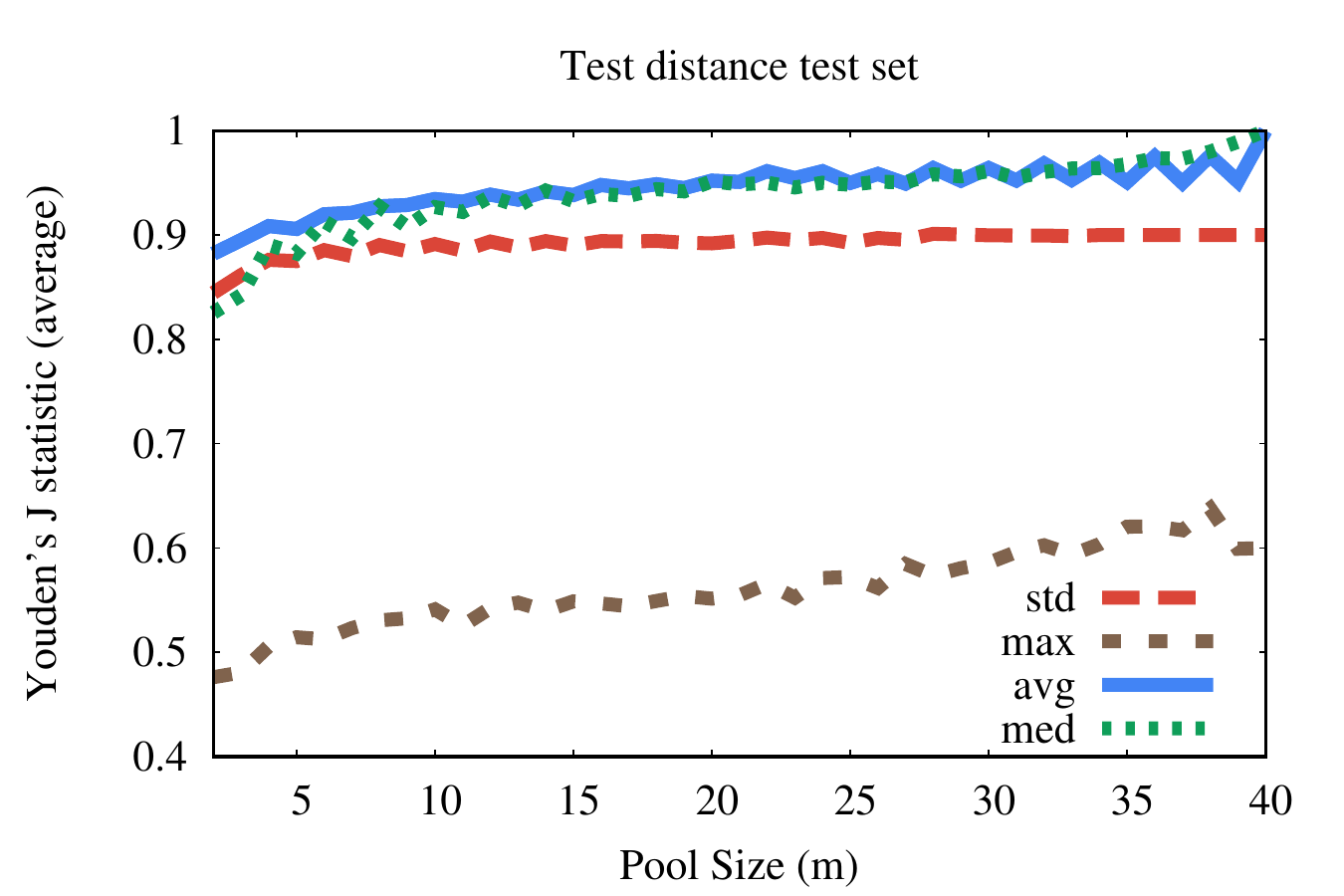}\hspace{-5mm}
\includegraphics[width=0.52\textwidth]{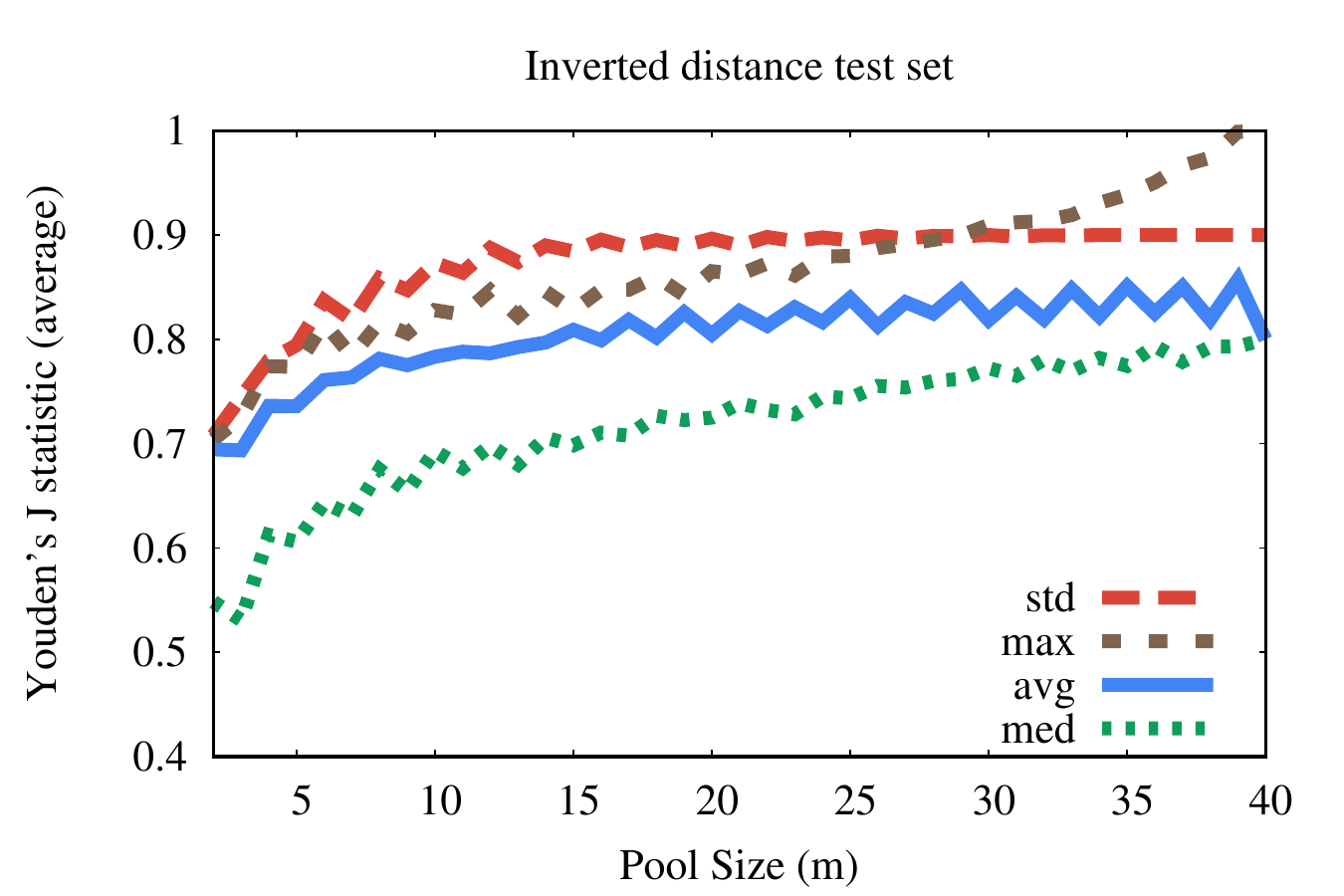}
\caption{Performance as a function of the pool size of the authority, for two
distance test sets (Test (left) and Inverted (right)) and different aggregations.}
\label{fig:extended}
\end{figure}

The results for $2 \leq m \leq 40$ are shown in~\cref{fig:extended},
using test and inverted images as distance test sets.
Note that there is only one possible training set when $m=40$, and due to this, and the size of the test set,
$J$ is a multiple of 0.1 in this case.

Increasing $m$ does not result in a significant gain in the best performance except when $m$
is small.

\subsection{Detecting an Adaptive Attack}
\label{sec:adaptive-attacks}

Although the provider can always decide not to insert a backdoor if it suspects
that a mystery shopping is taking place, 
it is necessary to consider the possibility that the provider is aware of
the detection method that the authority might use and it actively tries to avoid
the detection of the inserted backdoor.

Here, we assume that the provider knows the exact method that will be used for detection.
The goal of the provider is to poison the model in such a way that our detection method fails.
This can best be achieved if the sample-wise distance function we apply does not differentiate
between poisoned and clean models.

Let the sample-wise distance be CosL, the best performing choice so far.
We propose an adaptive attack against this distance function based on the idea of distillation~\cite{distillation},
a known approach for hiding backdoors~\cite{distillationdefense}.
First, the provider trains a clean model $\theta^*$, and then it trains a poisoned model $\tilde{\theta}$ using a loss function
that combines an accuracy objective and a distillation objective that acts on the logit layer.

Let $D_p=D \bigcup B$ be the poisoned dataset with $B$ containing the backdoor samples.
The loss function of a sample $(x,y)\in D_p$ on model $\theta$  is
\begin{equation}
\label{eq:loss_adaptive}
\ell((x,y),\theta) =\ \alpha\ \mbox{CrossEntropy}(f(x,\theta),y) +
\mathbb 1_{D}(x,y) (1 - \alpha)\ \mbox{CosL}(z(x,\theta^*),z(x,\theta)),
\end{equation}
where the second term is the distillation term minimizing sample-wise distance.
Parameter $\alpha\in [0,1]$ represents the tradeoff between cross-entropy and distillation.
The indicator function $\mathbb 1_{D}$ makes sure that only clean training samples use
distillation.

During this adaptive attack, the adversary initializes the model with the weights of its clean reference model.

\subsubsection{New model pools}

\begin{figure}
\hspace{-8mm}
\centering
\includegraphics[width=0.55\textwidth]{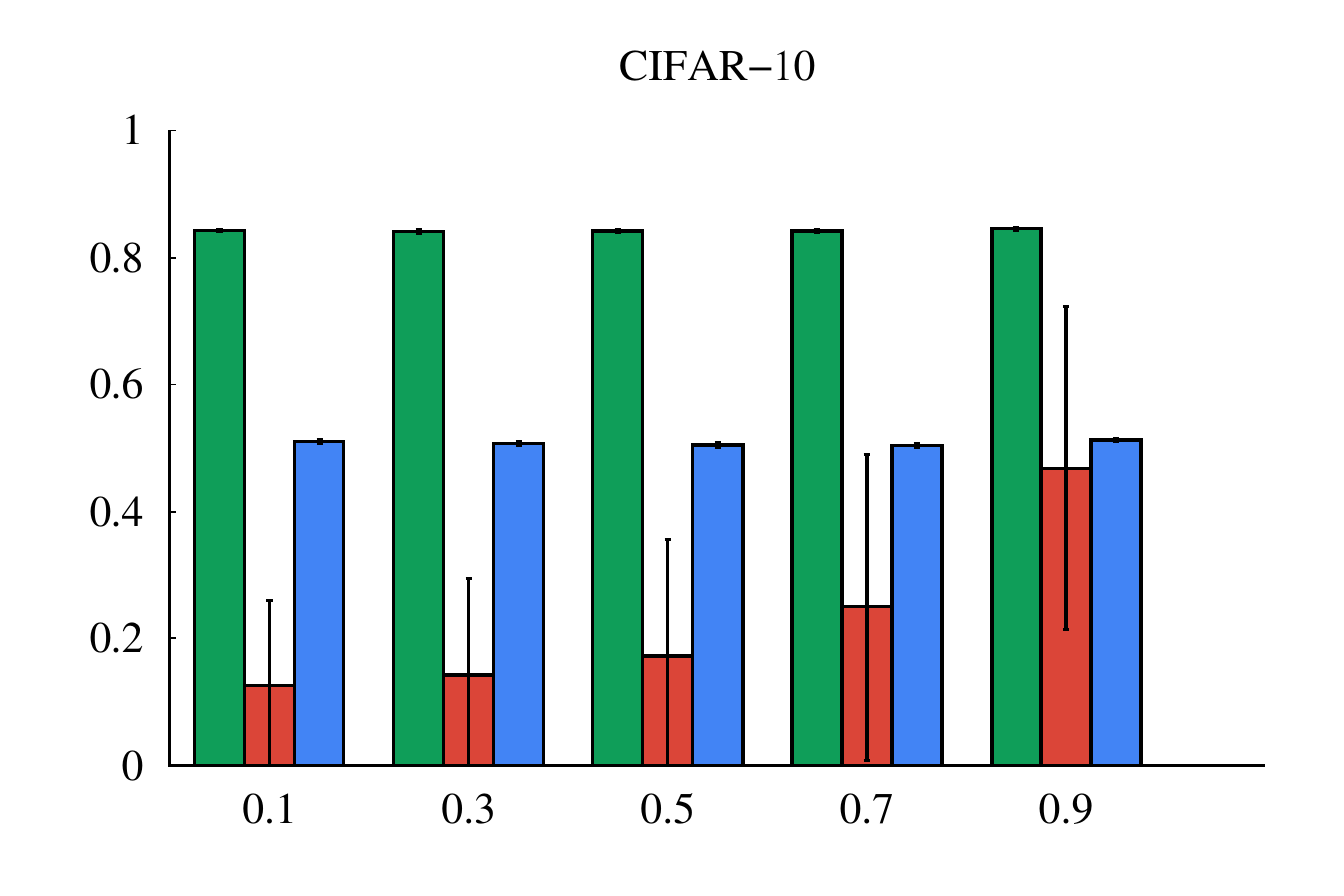}\hspace{-11mm}
\includegraphics[width=0.55\textwidth]{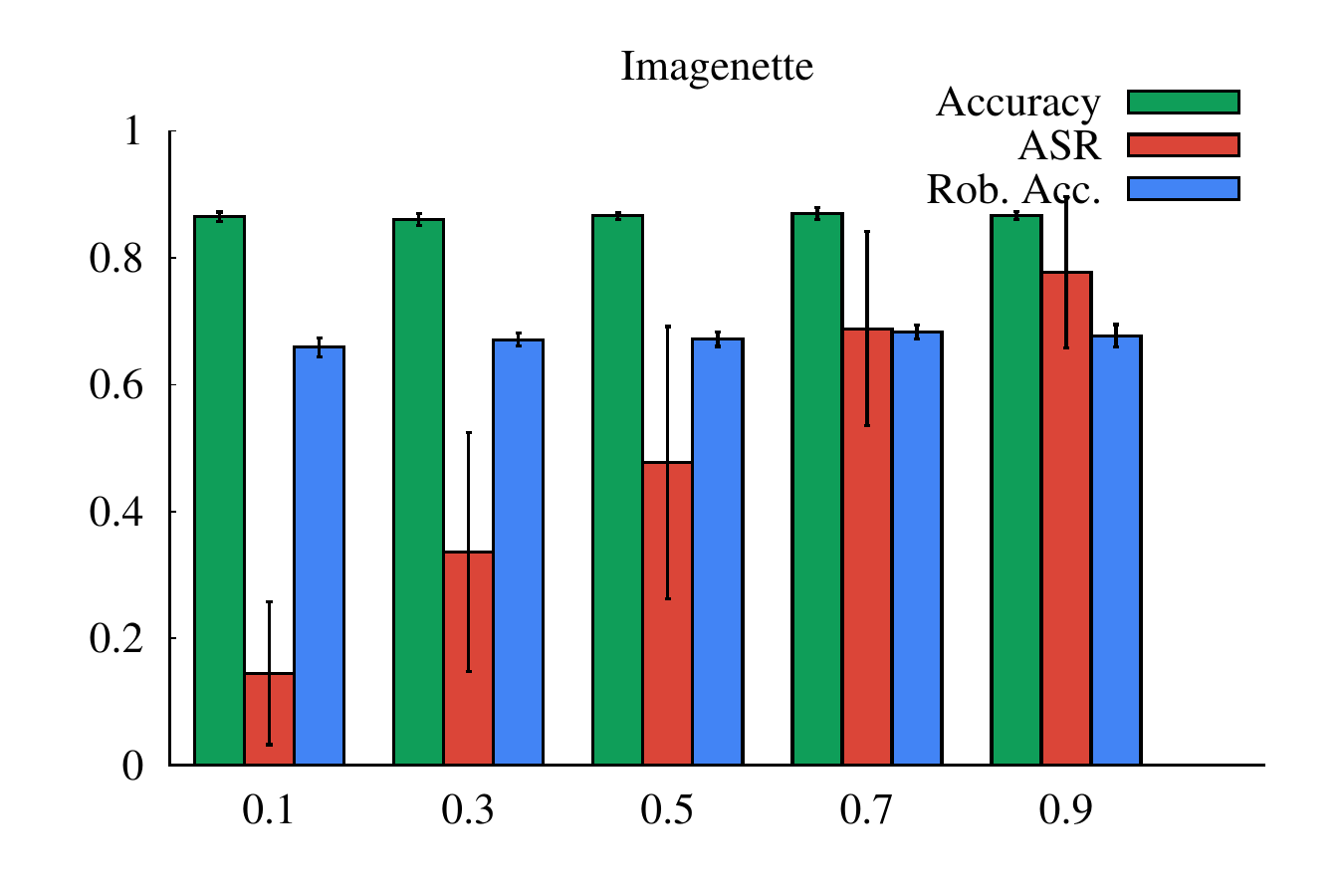}
\caption{Basic model statistics (accuracy, attack success rate (ASR), and robust accuracy) on the robust CIFAR-10 (left) and
robust Imagenette (right) adaptive model pools for different values of the $\alpha$ parameter.}
\label{fig:defpre_acc}
\end{figure}

\begin{figure}
\centering
\hspace{-3mm}\includegraphics[width=0.35\textwidth]{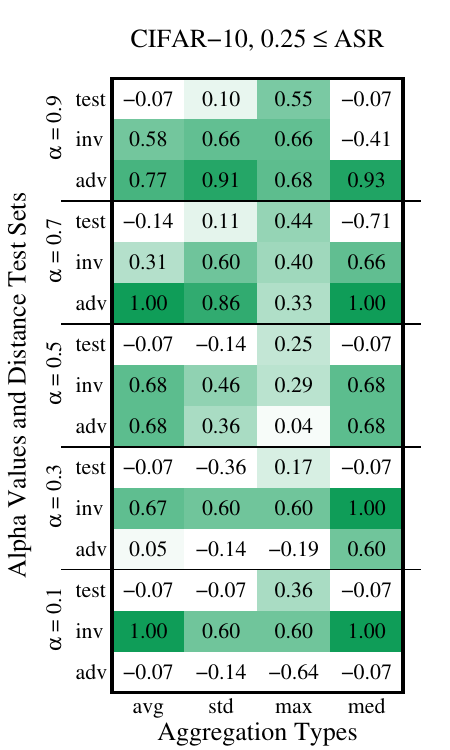}\hspace{-15mm}%
\includegraphics[width=0.35\textwidth]{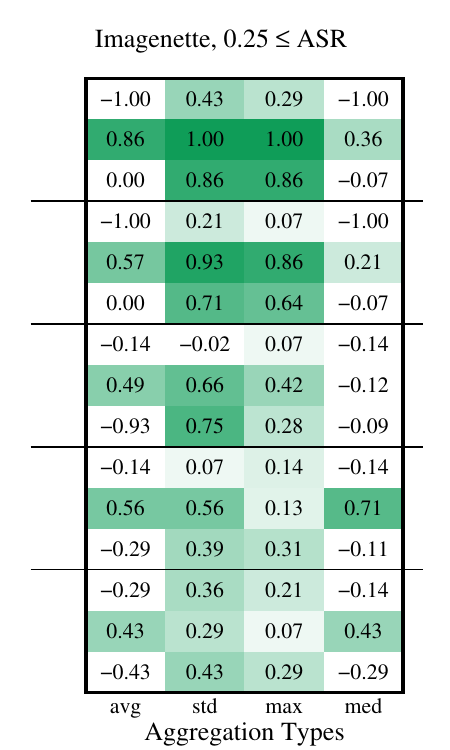}
\caption{Youden's J statistic with the CosL distance and the adaptive model pools for different $\alpha$ values, distance
test sets, and aggregations.}
\label{fig:minasr25_J_heatmap}
\end{figure}

To evaluate the adaptive poisoning attack, we created new model pools of 14 clean and
14 poisoned models.
More precisely, we created a new Imagenette robust ($\epsilon_\infty=4/255$) model pool
and a CIFAR-10 robust ($\epsilon_\infty=8/255$) model pool
for every $\alpha\in\{0.1, 0.3, 0.5, 0.7, 0.9\}$.
The pools with different values of $\alpha$ share the same 14 clean models with each other and their corresponding non-adaptive pool.
\Cref{fig:defpre_acc} depicts the basic properties of these pools.

\subsubsection{Evaluation of adaptive pools}

Let us first examine the J statistics obtained with the combinations of our distance test sets and aggregation types,
following the same methodology as used in~\cref{sec:evdesignspace}.
\Cref{fig:minasr25_J_heatmap} shows the results got just using the models with 0.25~$\leq$~ASR.
The inverted distance test set combined with the standard deviation aggregation method is the most stable option.

\begin{figure}
\centering
\hspace{-5mm}
\includegraphics[width=0.5\textwidth]{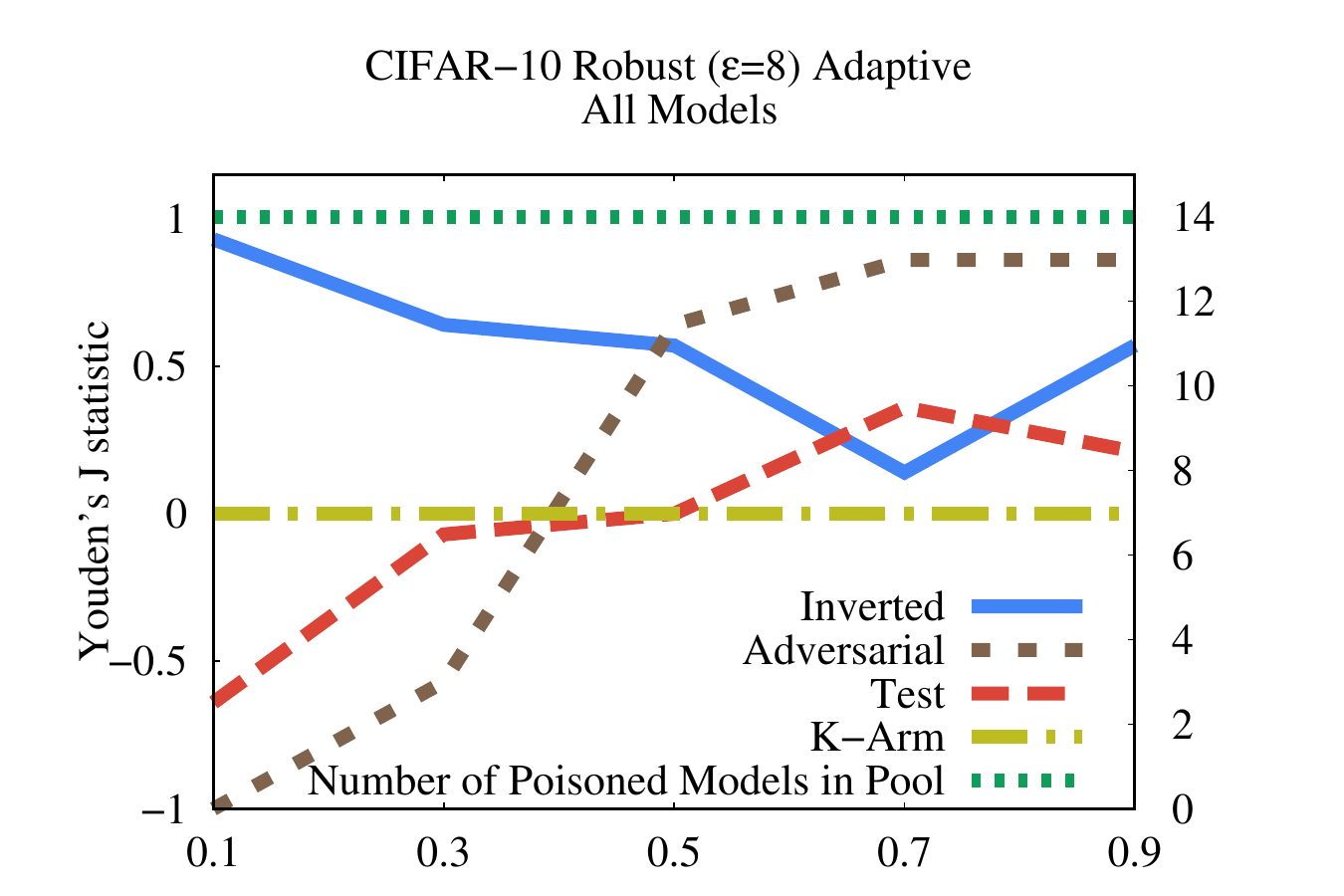}\hspace{-9mm}
\includegraphics[width=0.5\textwidth]{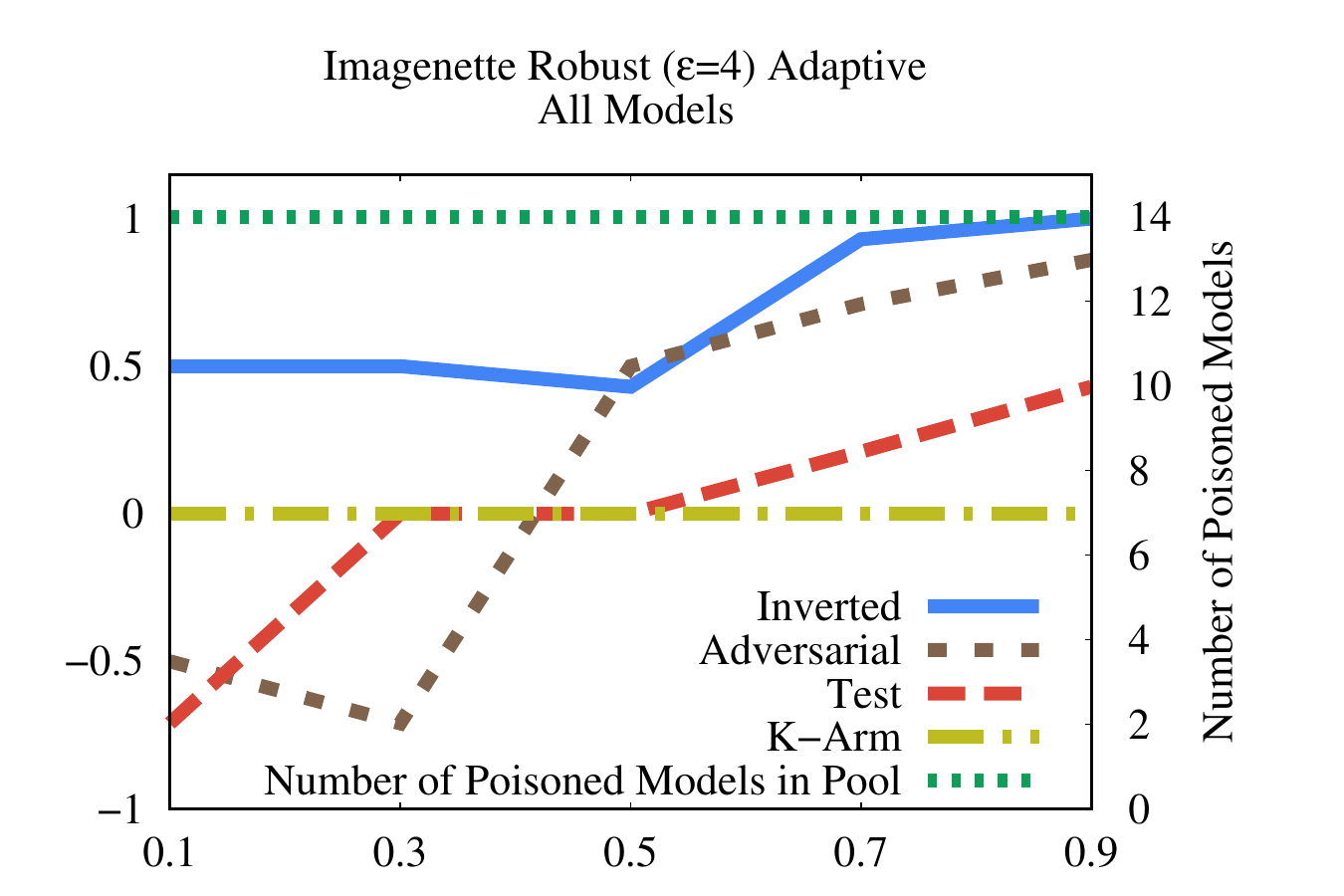}\\
\hspace{-5mm}
\includegraphics[width=0.5\textwidth]{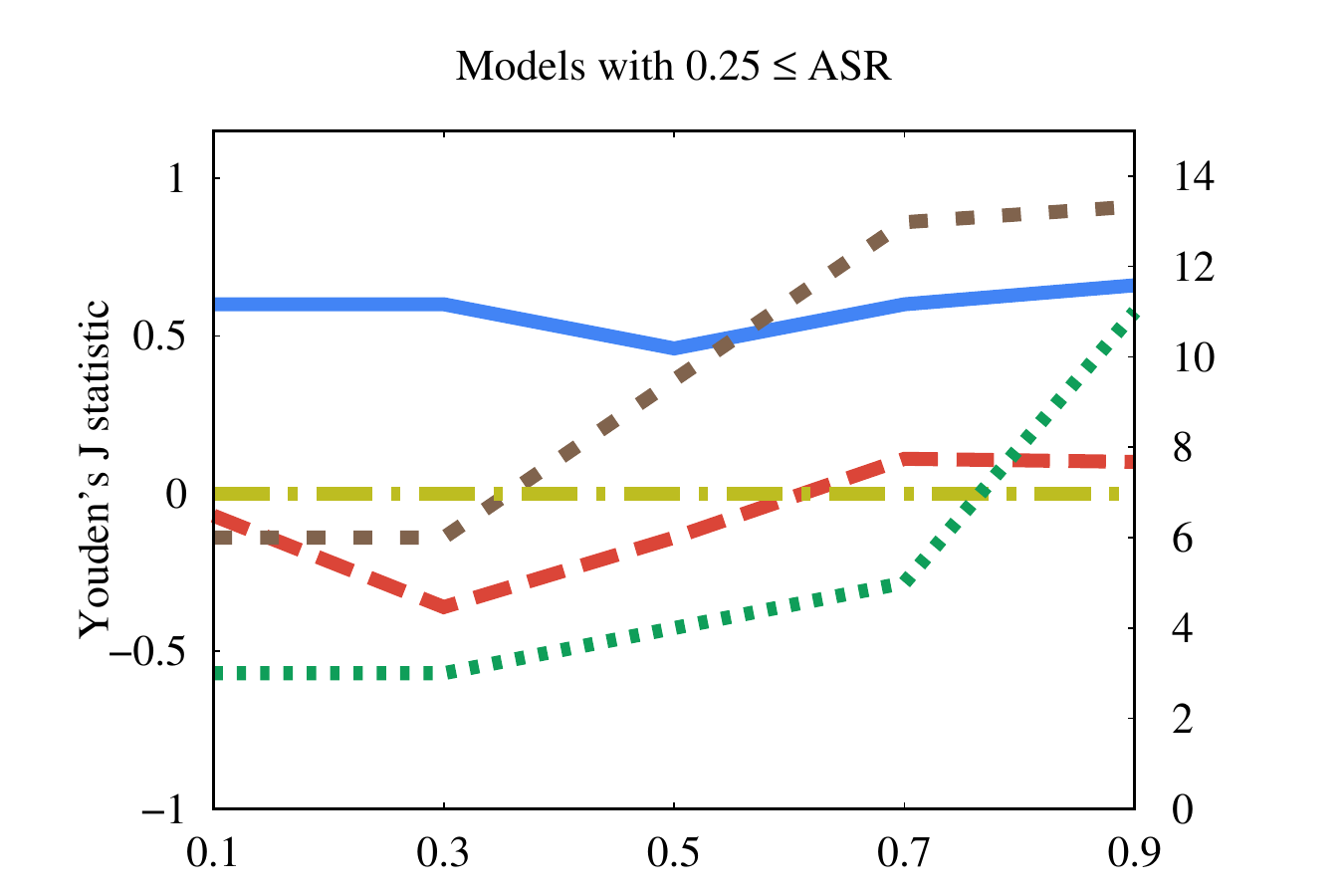}\hspace{-9mm}
\includegraphics[width=0.5\textwidth]{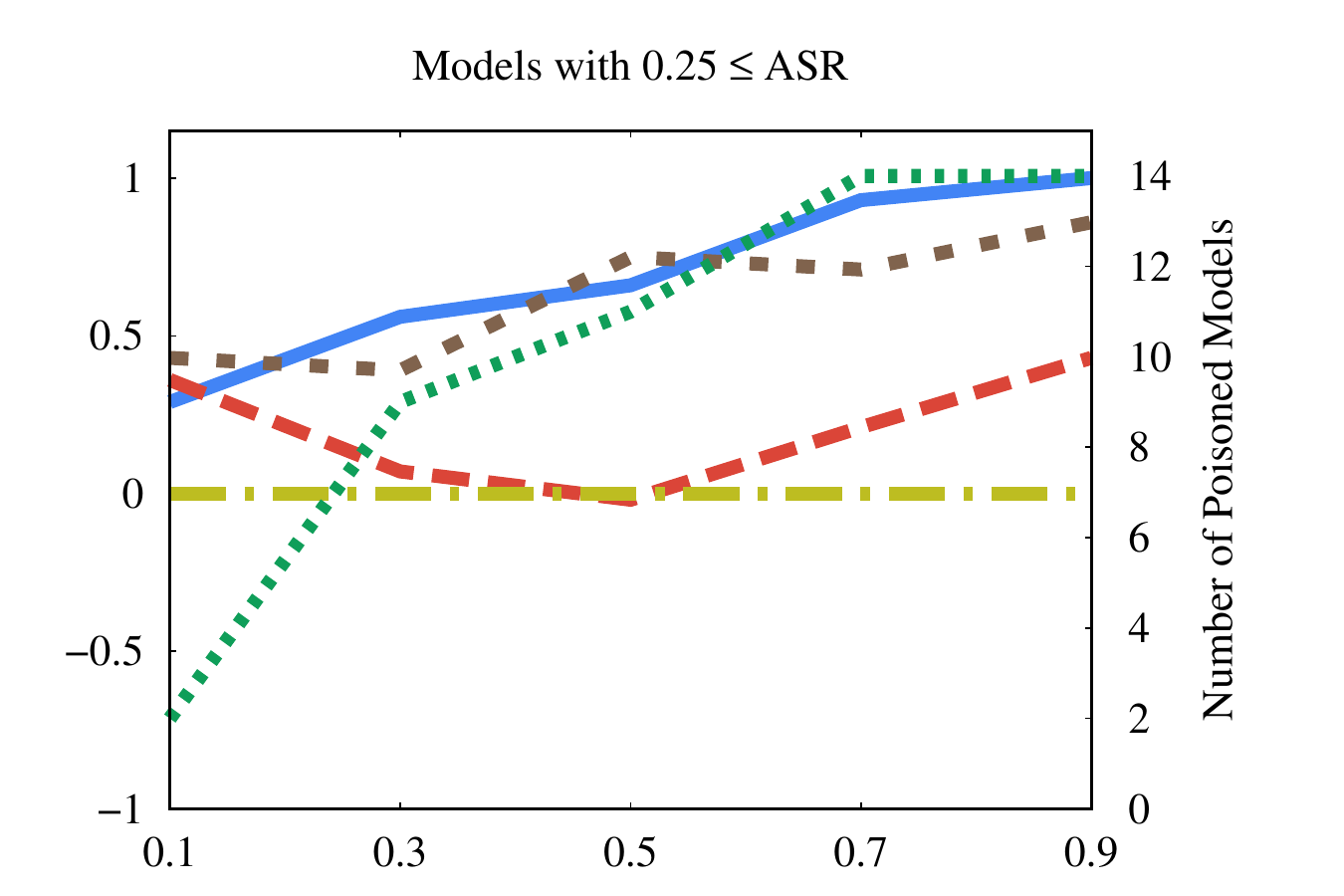}\\
\hspace{-5mm}
\includegraphics[width=0.5\textwidth]{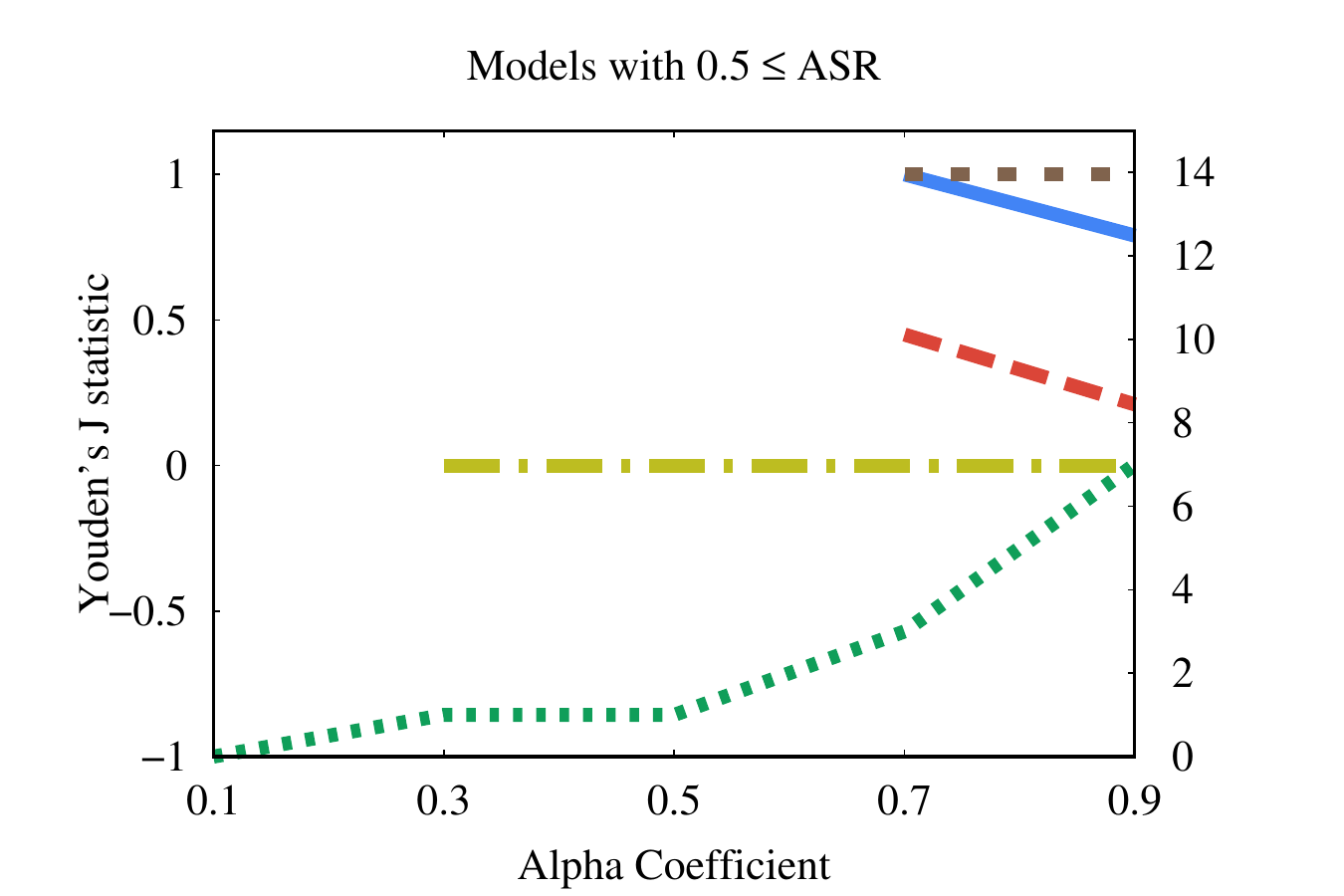}\hspace{-9mm}
\includegraphics[width=0.5\textwidth]{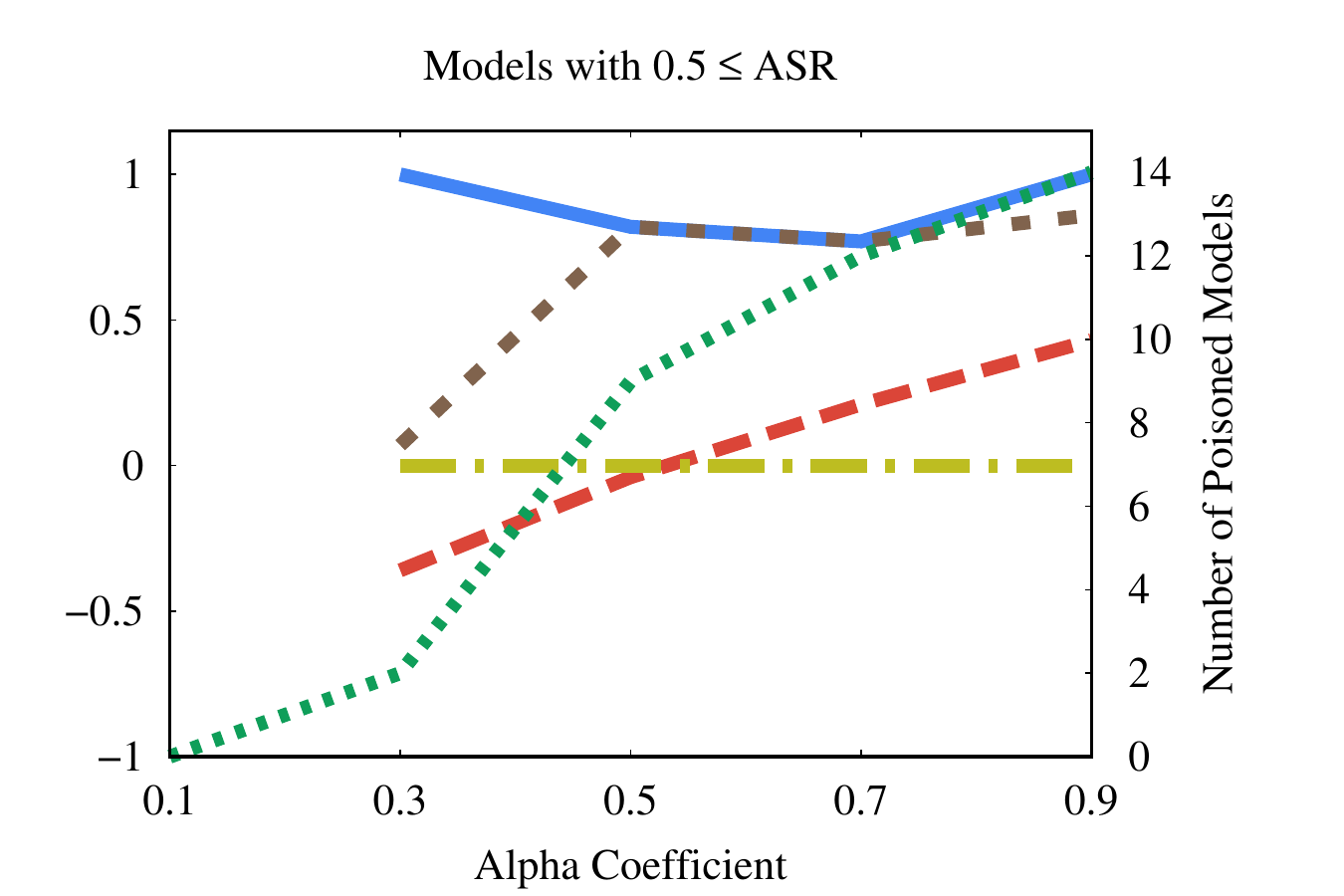}
\caption{Youden's J statistic for detecting the adaptive attack as a function of the $\alpha$ coefficient. The CosL distance
and standard deviation aggregation were used. The number of poisoned models is indicated on the right vertical axis.
Note that if the number of poisoned models is fewer than two then the evaluation of our method is not well-defined.}
\label{fig:defpre}
\end{figure}

\Cref{fig:defpre} shows the same data in a plot format, but with only the standard deviation aggregation,
and adding the case of unfiltered model pools (0~$\leq$~ASR) and pools with 0.5~$\leq$~ASR.
Here, with low values of $\alpha$, where the backdoor behavior is most hidden, the
inverted distance test set clearly outperforms the other alternatives for all the thresholds of ASR.
Also, for higher ASR thresholds the inverted set performs best overall.
Recall that models with a high ASR are interesting because these models are those
where the adaptive attack was the most successful.

\subsubsection{Ablations of Inverse Image Generation}
\label{sec:ablation}

Now that we have established that the inverted distance test set is the most promising approach
in the harder cases, let us perform an ablation study of this set generation method.
We do this by evaluating a number of variants of the method, namely 

\begin{itemize}
\item \emph{Without a prior}: we optimize the input image directly without using a prior generator network
\item \emph{Single step}: instead of the two-step method, we solve the problem in \cref{eq:loss_feat} by directly optimizing
the prior generator network in a single step
\item \emph{Invert only}: we simply invert the network using the prior generator network, but we perform no
other optimization
\end{itemize}

\begin{figure}
\centerline{\includegraphics[width=0.45\textwidth]{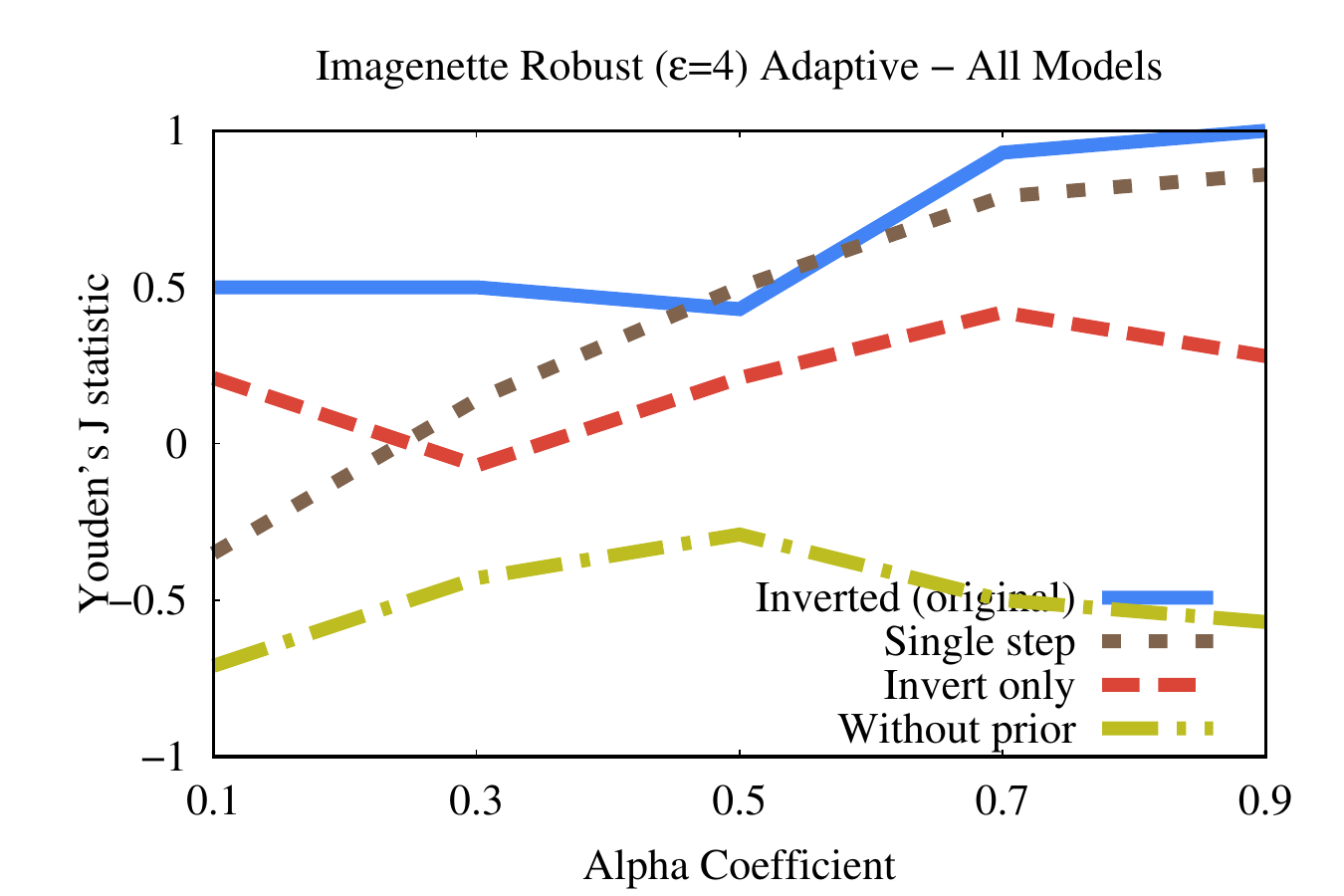}}
\caption{Ablations of Inverted distance test set generation.}
\label{fig:c10_defpre_generated}
\end{figure}

\Cref{fig:c10_defpre_generated} shows our results with the ablated versions of inverted distance test set generation.
Across almost all $\alpha$ values, the proposed method outperformed the simplified variants,
confirming that all the design elements contribute to the effectiveness of our approach.

\subsection{A Diverse Model Pool}
\label{subsec:diversepool}

Here, we evaluate the realistic case where the provider takes advantage of the degrees of freedom available to it to create
different kinds of backdoors.
To model this, we create a test model pool (to model the provider) that contains a diverse set of models, and we
create a larger training model pool (the pool of the authority) that also captures the diverse training strategies we
examined. Note that in our application scenario \emph{the authority controls the recipe of training}, and the provider
gains nothing by deviating from this, as such deviations only increase the chance of detection.
Therefore, the degree of
freedom available to the provider (apart from selecting the backdoor distribution) is whether it applies an adaptive attack strategy.

We evaluate this scenario on Imagenette, using robust ($\epsilon_\infty=4/255$) training.
Recall that the authority controls the type of training and the dataset as well, so this choice does not cause any loss of generality.
We start with the pool used in~\cref{subsec:num-of-models} which contains 50 clean and 50 poisoned models.
We include 40 clean and 40 poisoned randomly selected models in the training pool and the remaining models are
inserted into the test pool.

We extended the set of models using all the adaptive attack model pools we evaluated in~\cref{sec:adaptive-attacks}.
To be more precise, we included 50 adaptively poisoned models from these pools (selected from the models with 0.25~$\leq$~ASR) in the training
pool.
We trained 50 additional clean models and added these to the training pool as well.
Altogether, the training pool contained 90 poisoned models of several types and 90 clean models.

We also extended the test pool by training 5 adaptively poisoned models for every $\alpha\in\{0.5, 0.7, 0.9\}$, hence we added
15 models in total. 
We also trained 15 new clean models and added these to the test pool.
Note that these models all have 0.25~$\leq$~ASR.
This way, the test pool contained 25 poisoned and 25 clean models.

\begin{table}
\caption{Youden's J statistic on the diverse pool. The best J values in each column are in bold.}
\label{table:R18_imagenette_robust_extended_defpre_minASR025}
\begin{center}
\begin{tabular}{ l r r }
    Size of training model pool: & 28 & 180 \\ \hline
    Inverted & {\bf 0.68} & {\bf 0.72} \\
    Test & 0.39 & 0.48 \\\hline
    ULP~\cite{ULP-detector} & - & 0.28 \\
    K-ARM~\cite{k-arm-detector} & 0.00 & 0.00 \\
    Ex-Ray~\cite{exray-detector} (test data) & -0.04 & -0.04 \\
    Ex-Ray~\cite{exray-detector} (train data) & 0.08 & 0.08
\end{tabular}
\end{center}
\end{table}

We now present the results in \cref{table:R18_imagenette_robust_extended_defpre_minASR025}.
In the first column, a small pool of size 28 was sampled from the large pool of 180 models
independently (but with an equal number of clean and poisoned models)
500 times, and the average of the resulting 500 J values is included in the table.
In the second column, the full pool was used.
In both cases, the test pool was used for evaluation.

The best J that we got was with our method using the Inverted distance test set.

The large training pool allows us to test ULP~\cite{ULP-detector}, which is
based on training a meta-classifier over the models in the training pool and
so requires a large model pool.

We used the hyperparameters suggested by the authors.
Out of fairness, we should mention that ULP is based on the assumption that
the training backdoor models contain the same trigger pattern as the test models,
which is not the case in our scenario.

We also include results obtained with Ex-Ray~\cite{exray-detector} and K-Arm~\cite{k-arm-detector}, but we should add that these methods do not use training
at all, so the results shown here were simply calculated based on the test model pool.

\subsection{Generalization over Attack Types}

The provider has a limited degree of freedom because it must create models that appear very similar to those
that are requested by the authority.
Still, the adaptive attack represents a possible source of variability,
because we can set different values of $\alpha$.
Here, we discuss how robust the detection method is to this source of variability.
We do this by evaluating all the possible cases where the authority and the provider
use different attack methods.

\begin{figure}
\centering
\hspace{-2mm}\includegraphics[width=0.4\textwidth]{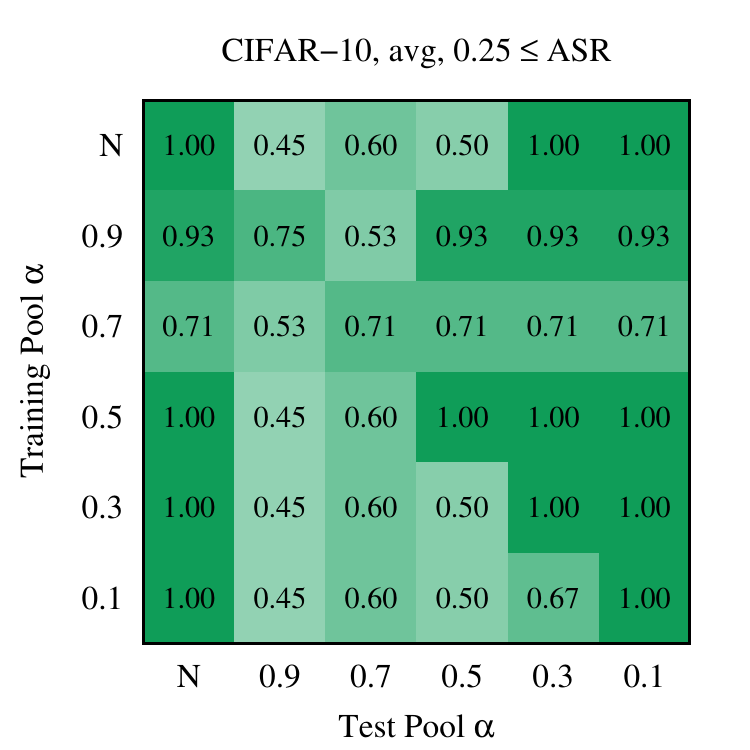}\hspace{-3mm}
\includegraphics[width=0.4\textwidth]{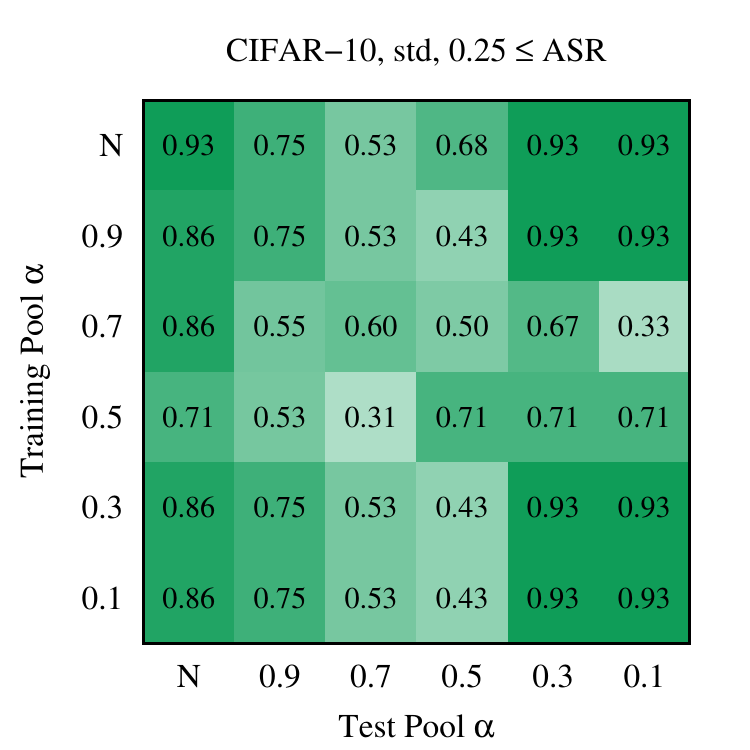}\\
\hspace{-2mm}\includegraphics[width=0.4\textwidth]{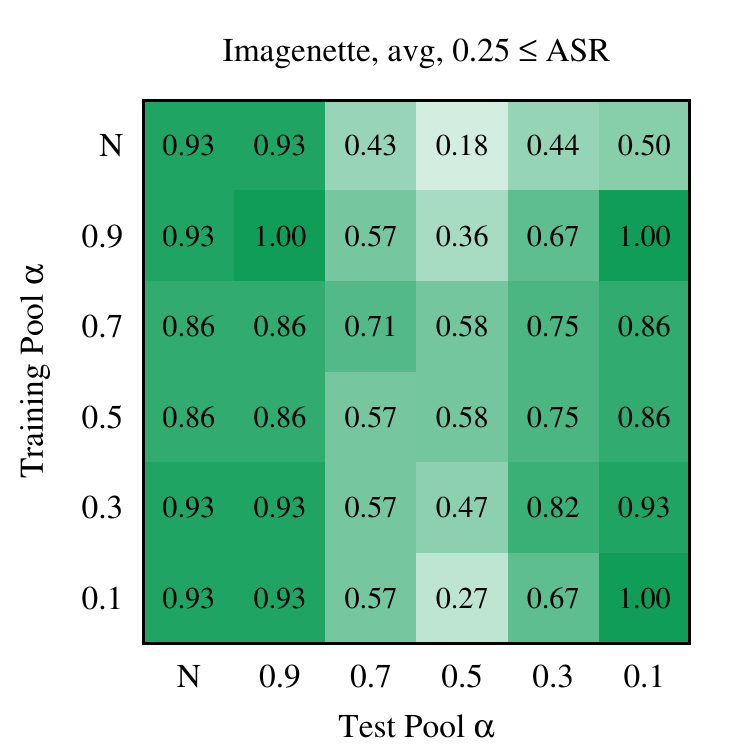}\hspace{-3mm}
\includegraphics[width=0.4\textwidth]{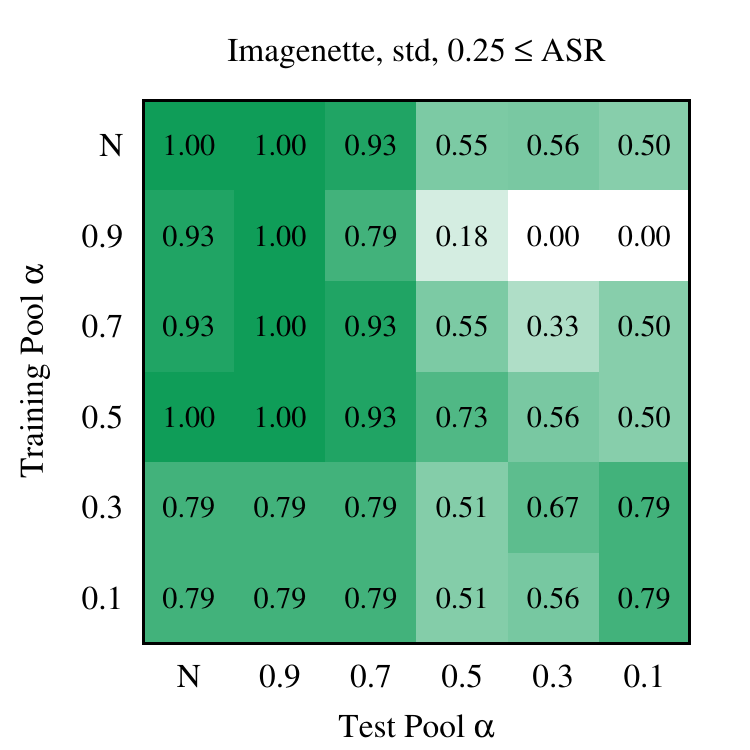}
\caption{Youden's J statistic for all possible combinations of attacks methods used by the authority (vertical axis)
and the provider (horizontal axis). In each case, the CosL distance was used on the
inverted distance test set. Two aggregation methods are included, namely average and standard deviation.
`N' denotes the non-adaptive robust pool.}
\label{fig:c10_mtx}
\end{figure}

\Cref{fig:c10_mtx} shows the J statistics with all such combinations.
We measure J with the help of the adaptive attack model pools described earlier.
In more detail, the optimal decision threshold is calculated using the authority's model pool (the
training pool) and J is calculated based on the classification of the models in the provider's
pool (the test pool) using this threshold.

The method is robust, as there are only a few cases where J is small. 
Most importantly, the \emph{non-adaptive (vanilla) training pool is robust} when using the inverted distance test set and
standard deviation aggregation.

\subsection{The Inverted Distance Test Set: Examples}
\label{subsec:examples}

\begin{figure*}
\centering
\hspace{-4mm}
\begin{tabular}{c}
0 \\[2.5mm]
1 \\[2.5mm]
2 \\[2.5mm]
3 \\[2.5mm]
4 \\[2.5mm]
5 \\[2.5mm]
6 \\[2.5mm]
\fbox{7} \\[2.5mm]
8 \\[2.5mm]
9 \\[-15mm]
\end{tabular}
\raisebox{-40mm}{\includegraphics[width=0.46\textwidth]{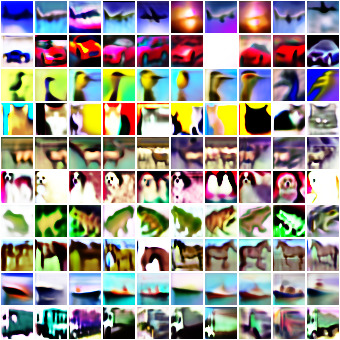}}\hfill
\raisebox{-40mm}{\includegraphics[width=0.46\textwidth]{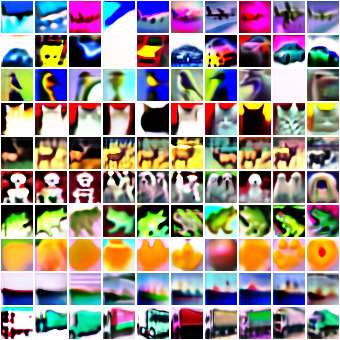}}
\caption{CIFAR-10 Inverted distance test set images for a clean robust model ($\epsilon=8/255$, left) and
for a poisoned robust model ($\epsilon=8/255$, right)).
Poisoned class: horse (7), backdoor class: fruit and vegetables  (CIFAR-100--4).
CIFAR-10 classes are listed top down, starting from class 0.}
\label{fig:cifar10inverted}
\end{figure*}

Motivated by the performance of the inverted distance test set, let us take a closer look at examples
of such sets on CIFAR-10.
\Cref{fig:cifar10inverted} shows two sets of generated inputs: one generated based on a clean model, and
one based on a poisoned model.
The poisoned class was the `horse' class,
and the backdoor OOD distribution was chosen to be the `fruits and vegetables' superclass of CIFAR-100.

This example demonstrates the intuition behind model inversion: our inversion algorithm is biased towards
\emph{realistic input samples that are different from clean samples}, so it creates fruit-like images that activate the `horse' class.
These generated backdoor inputs are very far from clean images.
This in turn results in a large sample-wise distance
between the clean and poisoned models over such generated inputs.

\begin{figure}
\centering
\rotatebox{90}{0.9\ \ \ 0.7\ \ \ 0.5\ \ \ 0.3\ \ \ 0.1}
\includegraphics[width=0.45\textwidth]{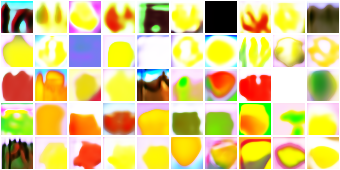}
\caption{CIFAR-10 inverted images for robust models ($\epsilon=8/255$) with different values
of $\alpha$ (adaptive attack parameter).
Poisoned class: horse (7), backdoor class: fruit and vegetables  (CIFAR-100--4).
Only the poisoned class is shown.}
\label{fig:cifar10_defpre-7-4}
\end{figure}

\Cref{fig:cifar10_defpre-7-4} shows the effect of the adaptive attack.
Recall that for small $\alpha$ values the adaptive attack is weaker and the backdoors
are harder to activate.
Indeed, for $\alpha=0.1$ the generated images are less fruit-like.

\section{Conclusions}

We studied a mystery shopping scenario where a consumer protection authority
tests whether a MLaaS provider inserts semantic backdoors during a training task.
In our method, the authority creates a small model pool containing clean and
poisoned models and calibrates a decision method based on this pool.

We showed through a large set of experiments that among the many options we examined,
the most robust one across the board is based on using adversarial training as a
mystery shopping task, and model distance is best measured using the CosL
distance (the cosine distance of the logit layers) over a set of samples that
are generated using model-inversion, biased towards generating backdoor inputs for
poisoned classes.
This method shows good performance even in detecting adaptive attacks by the provider.

\subsection{Practical Considerations}

Here, we share some thoughts on a number of practical issues related to
our research from the point of view of consumer protection authorities.
These include the cost of the methodology and some ethical and legal considerations.

\subsubsection{Cost}

Our method involves training model pools on a given problem, and our
best pools are based on adversarial training.
Due to these factors, the cost will be about two orders of magnitude higher than
that of training a single model.
Also, the authority needs to have access to expert knowledge in order to
design and create the pools.

On the other hand, this investment has to be made only once and 
the resulting detection thresholds will remain usable any number of times
for mystery shopping, as long as the testing task remains a secret.

\subsubsection{Legal and Ethical Issues}

Our results do not open any new legal or ethical issues within machine learning,
in fact, our scenario is less problematic than usual because \emph{the trained models are not
actually deployed}.
They remain private and unused.
However, the idea of using mystery shopping in this context is somewhat unusual so it is worth considering
some of the implications.

Since mystery shopping will most likely be executed by sub-contractors to hide
the identity of the authority, it is imperative that these sub-contractors are aware
of the nature of the task they are carrying out.
Also, it is advisable to publicly announce that the given authority executes
such mystery shopping projects for quality control and as a safety measure.

Obviously, the data involved in the project must comply with every data protection
regulation applicable to the given authority, despite the fact that the
trained models are never deployed.

\subsection{Limitations}

Let us now mention some of the limitations of our work.
Although we assumed that the provider uses a backdoor distribution that is unknown
and unpredictable to the authority, the provider can insert extra inputs that
might act as backdoors in many ways.
In other words, while the attacker is incentivized to include a backdoor that is OOD relative to the
clean distribution (otherwise no attack is necessary in the first place),
there is an open ended set of possibilities of such OOD distributions that we could
obviously not cover experimentally.

Also, evaluating the case of the adaptive attack is essential.
We presented a best-effort adaptive attack, but this is also only an empirical argument.
The possible set of adaptive attacks is also open-ended and we cannot guarantee that
a better attack will not be discovered that is even more successful in hiding the backdoor
from our detector.

We should mention here that in some of our experiments we used only two common datasets (CIFAR-10 and Imagenette),
and one network architecture (ResNet-18).
However, we included tests with additional datasets (VGGFace2 and ImageWoof) and architectures (ConvNeXt and ViT)
as well in some scenarios that suggest that the method is robust to these choices.
Also, it should be kept in mind that in our scenario
the authority controls what dataset and architecture is used.
This allows the authority to even optimize the dataset and the architecture to best support the
detection method.

Also worth mentioning that our framework is not suitable for testing services that are based on the provider's private data,
as that precludes the authority from creating the necessary reference pool.

While we provide extensive empirical evaluations to demonstrate the effectiveness of our approach,
we do not present mathematically rigorous theoretical explanations for our results.
The effect of poisoning on model distance is intuitive,
but not formally described beyond the metrics themselves.

Furthermore, poisoning is not the only factor that can increase the model distance:
for example, a dishonest provider might try to cut costs by training for fewer epochs than agreed upon,
or a careless provider might fail to pay attention to the instructions.
Our method could easily give a false positive result in these situations.
However, similarly to poisoning, such practices negatively impact the customers,
therefore their detection is not necessarily a flaw.

\section*{Acknowledgments}
This work was supported by the European Union project RRF-2.3.1-21-2022-00004 within the framework
of the Artificial Intelligence National Laboratory and project TKP2021-NVA-09, implemented with the
support provided by the Ministry of Culture and Innovation of Hungary from the National Research,
Development and Innovation Fund, financed under the TKP2021-NVA funding scheme.

\bibliographystyle{unsrtnat}
\bibliography{references,raw}

\end{document}